\documentclass[letterpaper]{article} 
\usepackage{aaai2026}  
\usepackage{times}  
\usepackage{helvet}  
\usepackage{courier}  
\usepackage[hyphens]{url}  
\usepackage{graphicx} 
\usepackage{natbib}  
\usepackage{caption} 
\usepackage{algorithm}
\usepackage{algorithmic}
\usepackage{xcolor}
\newcommand{\answerYes}[1]{\textcolor{blue}{#1}} 
 
\newcommand{\answerNA}[1]{\textcolor{gray}{#1}} 
 
\usepackage{newfloat}
\usepackage{listings}
\usepackage{graphicx} 
\usepackage{caption} 
\usepackage{subcaption} 
\DeclareCaptionStyle{ruled}{labelfont=normalfont,labelsep=colon,strut=off} 
\floatstyle{ruled}
\newfloat{listing}{tb}{lst}{}
\floatname{listing}{Listing}
\title{Posts of Peril: Detecting Information About Hazards in Text}
\author{ Keith Burghardt\textsuperscript{\rm 1},
 Daniel M.T. Fessler\textsuperscript{\rm 2,3,4},
 Chyna Tang \textsuperscript{\rm 2,5},
 Anne Pisor \textsuperscript{\rm 6},
 Kristina Lerman\textsuperscript{\rm 7},
 }

 \affiliations {
  \textsuperscript{\rm 1} School of Data Science, University of North Carolina at Charlotte, Charlotte, NC 28223, USA\\
 \textsuperscript{\rm 2}Department of Anthropology, University of California, Los Angeles, Los Angeles, CA 90095, USA\\
 \textsuperscript{\rm 3}Bedari Kindness Institute, University of California, Los Angeles, Los Angeles, CA 90095, USA\\
 \textsuperscript{\rm 4}Center for Behavior, Evolution, \& Culture, University of California, Los Angeles, Los Angeles, CA 90095, USA\\
 \textsuperscript{\rm 5}Department of Psychology, University of California, San Diego, San Diego, CA 92093, USA\\
 \textsuperscript{\rm 6}Department of Anthropology, The Pennsylvania State University, University Park, PA 16802, USA\\
\textsuperscript{\rm 7}Luddy School of Informatics, Indiana University, Bloomington, IN 47408, USA\\  
kburghar@charlotte.edu, dfessler@anthro.ucla.edu, pisor@psu.edu, chynst@g.ucla.edu, krlerman@iu.edu
 }

\begin{document}

\maketitle

\begin{abstract}
Socio-linguistic indicators of affectively-relevant phenomena, such as emotion or sentiment, are often extracted from text to better understand features of human-computer interactions, including on social media. However, an indicator that is often overlooked is the presence or absence of information concerning harms or hazards. Detecting such indicators in text is important because substantial research demonstrates that negative events are more likely to be attended to, and more likely to elicit a response. In addition, statements about hazards are often found to be more believable than statements about benefits. Here, we develop a new model to detect information concerning hazards, trained on a new collection of annotated X posts. We show that not only does this model perform well (outperforming, e.g., dictionary approaches), but that the hazard information it extracts is not strongly correlated with such widely used indicators as moral outrage, sentiment, and emotions. (That said, in accord with expectations, hazard information does correlate positively with such emotions as fear, and negatively with emotions like joy.) To demonstrate the utility of our tool, we apply it to two datasets of X posts that discuss important geopolitical events, namely the Israel-Hamas war and the 2022 French national election. In both cases, we find that hazard information, especially information concerning conflict, is common. We extract accounts associated with information campaigns from each data set to explore how information about hazards could be used to attempt to influence geopolitical events. We find that inorganic accounts representing the viewpoints of weaker sides in a conflict often discuss hazards to civilians, potentially as a way to elicit aid for the weaker side. Moreover, the rate at which these hazards are mentioned differs markedly from organic accounts, likely reflecting information operators' efforts to frame the given geopolitical event for strategic purposes. These results are first steps towards exploring hazards within an information warfare environment. The model is shared as a Python package to help researchers and journalists analyze hazard content. 
\end{abstract}

\begin{links}
\link{Code}{https://github.com/KeithBurghardt/HazardPackage}
\link{Datasets}{https://github.com/KeithBurghardt/HazardPackage/tree/main/ground_truth_data}
\end{links}

\section{Introduction}

Because the mind treats information regarding hazards differently than other types of information, developing techniques to identify hazard information can shed important light on the impact of online content. Humans pay particular attention to information that could signal potential hazards or dangers in their environment \cite{Soroka2019,ohman2001fears}. As a result, negative events tend to have more potent and lasting effects than their positive counterparts~\cite{baumeister2001bad,rozin2001negativity}. In a similar vein, negative news attracts more attention \cite{Soroka2019}, negative information spreads more readily than positive information \cite{bebbington2017sky,blaine2018origins,fay2021socially,ferrara2015quantifying,brady2017emotion}, and fear speech (negative information about a particular group) can drive groups towards conflict \cite{Saha2023}. 
Compounding overall negativity bias, people tend to find information about hazards more believable than information about benefits, a property termed negatively-biased credulity ~\cite{Fessler2014}. In accord with these features of negative information, such information is a factor in the spread of misinformation~\cite{vosoughi2018spread,ecker2022psychological,Youngblood2023}, suggesting that it could be a potent tool for information operations, i.e., sets of accounts aimed at influencing public opinion online. 

Although prior research has developed vital tools for identifying a variety of dimensions of social media content, to date, the importance of extracting information concerning hazards has been largely overlooked. We define hazards as negative events that entail a high probability of a significant (and often immediate) cost to individual interests~\cite{Fessler2014}. Although not all hazards are life-threatening or externally imposed (e.g., natural disasters such as hurricanes), they all represent threats that warrant attention and response. It  is particularly critical to analyze information concerning hazards during events such as terror attacks or wars~\cite{Choi2022,jamil2022detection,alshehri2020understanding}, where the dissemination of threat-related language can significantly influence public perception and behavior.

We address the gap in the existing literature by developing a multilingual model to recognize information concerning hazards (hereafter, for simplicity, ``hazards'') expressed in text messages on X (formerly Twitter). We validate the model on human-annotated data and show substantially higher performance over baselines. The model is highly scalable, enabling us to quantify the likelihood that hazards are present in millions of posts. We show that while the model's hazard classifier is somewhat correlated with negative emotions such as anger and fear, it goes beyond such indices, capturing additional salient information about potential dangers. 

We apply this model to two diverse multilingual X datasets that serve as case studies of the model's utility. Namely, we detect hazards in posts from 2023 about the Israel-Hamas war, and posts from 2022 about the French national election. The datasets are unique, from different locations and types of conflict (war versus political contest), as well as different languages, including Arabic, French, and Chinese. Moreover, these data contain material that we attribute to information operations.
We see notable differences in how information operation accounts describe threats (for example, emphasizing harms in Gaza and not in Israel), while also using hazards to demean an opponent. We also find that the prevalence or frequency of specific hazards differs greatly between different information operations, reflecting different interest factions within each dataset. These findings appear to show that hazards to civilians are consistently used to justify aid to the weaker side in a conflict. More broadly, these results demonstrate how identifying hazards can reveal salient features, potentially unmasking information operation tactics. 
Because our hazard-detection model has broad utility, we have created a package that enables any researcher to apply it to their own datasets; we do so recognizing that the present effort is a first step towards comprehensive hazard detection, as additional research will improve the model.

In summary, our contributions are as follows.
\begin{itemize}
    \item We develop and share a new model to extract hazards from social media posts.    
    \item To demonstrate the model's utility, we apply it to two newly collected online datasets, and contrast hazards to alternative indicators of affect and threat.
    \item We analyze hazards used in information operations to further reveal influence tactics.
    \item We share hazard-annotated data from a wide range of X posts so as to enable researchers to iteratively improve upon this model.
\end{itemize}

Overall, these results demonstrate the need for, and utility of, our new tool to detect information about hazards, especially in social media environments. 

\section{Related Work}

\subsection{Information Operations}
Our work focuses on hazards in information operations, coordinated efforts to change behaviors or opinions \cite{burghardt2018quantifying}, especially during important  events, such as the Israel-Hamas war \cite{dey2024coordinated}, or elections \cite{badawy2018analyzing,burghardt2018quantifying}. Detecting information operations is difficult, as, with few exceptions (e.g., \cite{luceri2023unmasking}), investigators generally lack access to ground-truth datasets with which to train or evaluate a model. Instead, researchers typically aim to detect strong coordination, such as unusually similar text or hashtags \cite{burghardt2018quantifying,luceri2023unmasking}, or quick reposts \cite{mazza2019rtbust}. In the present study, we use hashtag matching \cite{burghardt2018quantifying,luceri2023unmasking}, as well as a method by Luceri et al. \cite{luceri2023unmasking} which combines several different approaches together: co-posting, co-URL sharing, fast re-posting, hashtag matching, and text similarity. It then finds the eigenvector centrality of the network, as high eigenvector central nodes are more likely to be part of an information campaign.

\subsection{Linguistic Indicators in Text}
Detecting indicators from text has a long history. One of the first methods to do this was The General Inquirer \cite{stone1966general}, and, more recently, text indicators have been analyzed with LIWC \cite{pennebaker2001linguistic}. These dictionary-based approaches have subsequently been largely replaced by rule-and-word approaches, such as VADER \cite{hutto2014vader}, which have themselves been superseded by text embedding approaches, including SeerNet \cite{duppada2018seernet}, a moral outrage classifier \cite{Brady2021}, and DeepMoji \cite{felbo2017using}. The advantages of these approaches are that embeddings can learn how semantically similar content has similar indicator values (e.g., ``I am sad'' and ``This day was awful'' would be viewed as sad statements even when the two sentences do not have any words in common).  These approaches are based on GRU \cite{dey2017gate} or LSTM \cite{hochreiter1997long}; however, more recently, transformer-based models \cite{vaswani2017attention}, such as BERT \cite{devlin2018bert} and Sentence-BERT \cite{reimers2019sentence}, have become popular for uses such as detecting hate speech \cite{davani2022dealing} or emotions \cite{acheampong2021transformer}. Unlike prior approaches, transformer models account for context. Finally, these methods have been improved with SpanEmo \cite{alhuzali2021spanemo} or Demux \cite{Chochlakis2023}, which account for correlations between labels.

\subsection{LLM Approaches to Text Analysis}
While traditional methods to detect linguistic indicators are either dictionary approaches, e.g., LIWC \cite{pennebaker2001linguistic}, or supervised methods on smaller language models, such as Demux~\cite{Chochlakis2023}, there has been a recent advance in LLM appropaches, especially given their wide applicability \cite{cao2025toward}. These approaches include using Large Language Models (LLMs) to similarly detect emotion \cite{peng2024customising} or othering language \cite{gerard2025fear} via LoRA fine-tuning \cite{hu2022lora} and Prompt Tuning \cite{liu2021p}. These techniques (along with QLoRA \cite{dettmers2023qlora}) allow for efficient fine-tuning of multi-billion parameter models by adding a low-rank matrix to each layer. Alternative methods include few-shot learning \cite{Hong2025}, in which prompts are injected with additional examples to give context to a prompt. In contrast, reasoning is found to improve model performance, and therefore having LLMs explain their step-by-step reasoning improves their performance across a range of tasks \cite{Hama2024}.

Here, we develop a transformer-based model to extract hazards--a previously under-studied indicator--from social media posts. We apply a range of models to sentence embeddings and compare these with both LLM (GPT-3.5, GPT-4, and GPT-5 \cite{achiam2023gpt}) and dictionary approaches to detect hazards in text \cite{Choi2022}. As effective baselines, we include both few-shot prompting and reasoning, but leave a fine-tuned LLM baseline as future work. While Choi et al.'s analysis of threats in text is perhaps most similar to our project, we detect hazards rather than threats. Moreover, we apply multi-lingual AI models to detect hazards, which we find is a significant improvement over dictionary approaches, such as that of Choi et al.~\cite{Choi2022}. This work also compliments prior work detecting hazardous events in a time series \cite{jamil2022detection}, as well as explicit threats (dangerous speech) \cite{alshehri2020understanding}.
 
\subsection{Negatively-Biased Credulity} 
As noted above, a growing corpus of research documents that negative information spreads more readily than positive information. For example, negative content is more likely to be shared \cite{Martel2020,Youngblood2023}; moral-emotional language (often in large part negative in nature) spreads the most in partisan discussions \cite{brady2017emotion}; negative sentiment posts spread faster \cite{ferrara2015quantifying}; and moral outrage makes posts more viral \cite{Brady2021}. One factor likely contributing to the asymmetry in the spread of negative versus positive information is that negative information -- particularly information about potential dangers -- is more likely to be believed. 
For people past and present, believing false information about hazards has, likely been less costly on average than rejecting true information about hazards, since taking unnecessary precautions generally entails less dire consequences than does failing to take necessary precautions against threats; in contrast, there likely has not been an overarching pattern with regard to the respective costs and benefits of believing or not believing information about benefits. People therefore broadly find information about hazards more believable than information concerning benefits, a pattern termed negatively-biased credulity \cite{Fessler2014,Fessler2015,fessler2019believing,Samore2018,forgas2019role}. Of particular relevance in regard to information operations, work on negatively-biased credulity articulates with investigations on the role of credulity in recipients' susceptibility to manipulation \cite{Little2017,Kartik2007}. 

Although there are many points of contact between psychological research on negatively-biased credulity and existing work on the spread of information, misinformation, and disinformation, this construct also calls attention to potentially important distinctions that have been overlooked in prior research on information spread. With a logic grounded in evolutionary theory, a core insight of work on negatively-biased credulity is that not all negative information confers a similar survival advantage. Specifically, swiftly recognizing imminent threats ~\cite{ohman2001fears} (marked, for example, by fear) is more critical to survival than is revisiting past losses (marked, for example, by sadness). Hence, while negative emotions as a category may reduce belief in a false claim \cite{phillips2024emotional}, discussion of hazards seems to increase it \cite{fessler2019believing}. Likewise, although other evidence shows that overall emotion perception is associated with both poor misinformation discernment and misinformation sharing \cite{bago2021emotion,Martel2020}, such work does not differentiate between emotions attending messages concerning hazards and those attending messages concerning benefits. As a metric of content, it is therefore critical to distinguish information about hazards from negative valence in general, the presence of emotion-laden information, or other features of language. 

We expand on previous research regarding negativity by developing a novel hazard detection tool. We demonstrate this tool's utility by applying it to millions of social media posts. We employ large datasets that provide enough statistical power for us to assess how statements about hazards vary after major events, how they relate to other indicators, and how diverse groups discuss hazards. In turn, this can inform hypotheses about how hazard information is harnessed when trying to influence social media users.

\section{Research Methods}
All data collected and analyzed were determined to be exempt from assessment by the institutional review board of the lead author's university, where all modeling, data collection and data analysis were conducted. All annotations were non-human-subject research. In addition, all data were anonymized prior to analysis or annotation to minimize privacy risks. This was accomplished via anonymization of user names and profile content.

\subsection{Hazards Benchmark: Data and Model}
We curate a ground truth dataset with which to train models to recognize hazards. This process involves collecting and annotating posts for the presence of hazards. This ground truth dataset is then used to train a language model to classify hazards posts. 

\paragraph{Ground Truth Data}
To create the benchmark X post dataset, we first extracted 1,338 X posts containing at least one word from the Threat Dictionary \cite{Choi2022}. Although we are not aware of any dataset annotated for hazards, we chose this size as approximately the median number of posts annotated for valence across several previous datasets (cf. Table 1 in \cite{mendes2023quantifying}). We confirm that this is a sufficient sample through the model performance described in the Results section. In order to produce a representative sample, data are collected via X's Academic API between March 2006 (when X, then Twitter, was founded) and late 2022. 

We randomize the order of these posts and recruit Cloud Research annotators to label any hazards present therein via a Qualtrics survey. For each post, workers answer, ``Does the tweet describe a hazard (something that could impose harm or other costs on the author of the tweet or on others)?''. (Our annotations predate X's name change from Twitter, hence our use of ``tweet'' in the annotation question rather than ``post''.) Workers are paid \$2 for each assignment in which they annotate 10 random posts (on average this took 11 minutes to complete, hence compensation was equivalent to \$12 an hour). To account for workers who do not meaningfully complete the assignment, we add an easy question (specifically ``choose the answer to 2+2'') midway through the survey and remove any workers who did not complete it. We also remove annotations that are unfinished or take less than 200 seconds to complete (this was an arbitrary cutoff to better ensure that the annotations were not completed carelessly). As all data are annotated before 2023, we believe that the prevalence of workers annotating using LLMs \cite{veselovsky2023artificial} was minimal. 

To check inter-rater reliability, we used the R library \texttt{irrNA} \cite{irrna}, which assumes randomly missing data (this assumption is consistent with our annotation methodology, as we assign 10 posts at random to each rater). Applied to the dataset, we have an Intraclass Correlation Coefficient of: ICC(1) $=$ 0.12, ICC(k) $=$ 0.29, agreement of ICC(A,1) $=$ 0.17, ICC(A,k) $=$ 0.37, ICC(C,1)$=$0.17, and ICC(C,k) $=$0.37 (p-values $<0.001$). These values can be interpreted as moderate consistency but poor agreement. However, these values are in keeping with other subjective text rating tasks, including hate speech \cite{sachdeva-etal-2022-measuring} and emotions \cite{mohammad2018semeval}. The low ICC may also reflect the difficult nature of the task that we gave annotators, as they were trying to distinguish between posts with versus without hazards despite the fact that all of the annotated posts contain Threat Dictionary words.

All posts annotated by fewer than two crowd workers are discarded, resulting in a dataset with 1,131 posts for training, validation, and testing. We use Python's demoji library (https://pypi.org/project/demoji/) to convert all emojis to words in order to reduce artifacts in the embeddings. All data are collected according to the FAIR principles:
Findable (these data are directly available via the repository link at the end of the Introduction section, and contain a unique identifier with all metadata described), Accessible (the link is accessible to anyone), Interoperable (metadata use a formal and broadly applicable language), and Reusable  (all data are described in detail). In the repository link, we also include a datasheet for the dataset \citet{gebru2021datasheets}. 

\paragraph{Additional Dataset Annotations}
Random posts that we use to train the model might not be representative of the Israel-Hamas war and 2022 French election datasets to which the model is to be applied. To {address this potential limitation, we trained three students to annotate 150 random posts within each of the respective datasets. To adequately compare our model against the Threat Dictionary, and because these students were native English speakers, we translated all text to English prior to annotation. We specifically used X translated text for French posts, and translate Arabic text to English using Google Translate as the Twitter translations were not available for that set of text.  We provided each student with the same guidelines given to Cloud Research workers, but to ensure data quality, we also had students label posts ``-1'' if there was not enough information to determine whether the post contained a hazard (e.g., a post containing a URL), and ``1/2'' if the annotator had low confidence that the post contains a hazard due to an ambiguous scenario (e.g., parading around a body). To clean data, we removed any post that any of the annotators rated as ``-1'';  for the rare cases in which a post was labeled ``1/2'', we designated the post as ``not hazard''. This makes it especially difficult for the model, as it was trained on tweets containing hazards, and would presumably have a higher false-positive rate. After data cleaning, we have 124 annotated posts in the Israel-Hamas war dataset, and 135 posts in the 2022 French election dataset. The Fleiss Kappa scores for these annotations are 0.40 and 0.49 for the Israel-Hamas and French election datasets, respectively, representing moderate agreement. There were 74\% and 10\% of data labeled hazards in the respective datasets (which intuitively indicates a high rate of hazard discussions within a war context).
\paragraph{Model Training}
We use 90\% of data for training or validation and 10\% for testing, ensuring a sufficient amount of training data for each model. We choose from four multi-lingual sentence transformers: \texttt{distiluse-base-multilingual-cased-v2}, \texttt{paraphrase-multilingual-MiniLM-L12-v2}, \texttt{Qwen3-Embedding-0.6B}, and \texttt{stsb-xlm-r-multilingual} to embed text in the datasets. Multilingual text embedding models were used because of the different languages in each social media dataset. We then apply several supervised models to these embeddings, including XGBoost \cite{chen2016xgboost} (via \url{https://xgboost.readthedocs.io/}), neural networks \cite{tensorflow2015-whitepaper}, random forest \cite{RF,ho1995random,scikit-learn}, and support vector machines (SVMs) \cite{SVM,hearst1998support,scikit-learn}; the latter two are trained via \texttt{scikit-learn} \cite{scikit-learn}. We use BayesSearchCV to optimize hyperparameters for random forest, SVM, and XGBoost \cite{skopt}. We find the optimal parameters via five-fold cross-validation of the data split after ten iterations and minimize each model's default scoring metric. We also use GPyOpt Bayesian Optimization for a feedforward neural network \cite{gpyopt2016}. The best hyperparameters are shown in the GitHub repository.

We also experimented with augmenting human annotated posts with 5,000 GPT-3.5 annotated posts that contained words from the Threat Dictionary \cite{Choi2022}, and 5,000 posts collected at random (containing popular English keywords, namely any of the top 100 lemmas within a large corpus (\url{https://www.wordfrequency.info/samples.asp}) 
that X does not consider stop words. This augmentation did not significantly change the performance of the model. All models are trained via NVIDIA Tesla K80 GPU with 12GB of VRAM; we predict text via GeForce RTX 2080 GPU with 8GB VRAM. For comparison, we also use GPT-3.5, GPT-4, and GPT-5 \cite{achiam2023gpt} with few-shot learning and chain-of-thought prompting \cite{wei2022chain} via the prompt, ``Does the tweet [story] describe a hazard (something that could impose harm or other costs on the author of the tweet or on others)? Please answer `yes' or `no' and explain your thought process.'' and include zero, two, or five examples of posts containing hazards and posts that do not. All texts with ``yes'' are labeled ``hazard,'' and those without are labeled ``not hazard'' (rare situations where the LLMs do not know if a hazard exists in text are labeled ``not hazard'').

\subsection{Comparison to Alternative Text Indicators}
We compare our hazard detection model to alternative text indicators of affect-related phenomena: moral outrage \cite{Brady2021}, sentiment (VADER) \cite{hutto2014vader}, emotion detection (Demux) \cite{Chochlakis2023}, and threat words \cite{Choi2022}. These are state-of-the-art methods for detecting each indicator. These code are under a MIT license (VADER and Demux), and Creative Commons Attribution-NonCommercial-ShareAlike 2.0 license (moral outrage), respectively, and are adapted as needed to run on a GeForce RTX 2080 GPU with 8GB VRAM. All model outputs that we analyze are continuous (such as confidence values for emotions), with the exception of the Threat Dictionary \cite{Choi2022}, in which we indicate if a threat word is (1) or is not (0) present in a post. Because the Threat Dictionary is not necessarily a comprehensive dataset, we also found the top 3 nearest synonyms to each word in the Threat Dictionary using a pruned Word2Vec trained on the Google News Dataset \cite{bird-loper-2004-nltk}. We then compare the performance of this enhanced dictionary on annotated datasets to the best-performing model. We also compare how the enhanced threat dictionary results compare on the two datasets studied. The enhanced dictionary is available on our GitHub.

\subsection{X Data}
To demonstrate the hazard detection model's utility,  as case studies, we apply it to two separate datasets.
\paragraph{Israel-Hamas War}
Our analysis uses a corpus of 3.6M posts, from 1.3M accounts, spanning the period from August 31 to November 1, 2023, about the 2023 Hamas attack on Israel and Israel's subsequent invasion of Gaza. The posts were collected by querying X with a set of keywords related to the war: e.g., ``Israel,'' ``Hamas,'' ``Gaza,'' etc. These data are multilingual, with approximately 93\% of the posts in English, 6.5\% in Arabic, and a small proportion in other languages. 

\paragraph{2022 French Election}
We also analyze 5.9M posts, from 677K accounts on X, related to the 2022 French election that took place April 10--24, 2022. The dataset spans February 14 to June 30, 2022. The posts were collected by querying X with a set of keywords related to the election, e.g., major candidates such as ``Macron'', ``Marine Le Pen'', ``M\'elenchon'', etc. These data were also multilingual, with 95\% in French and 5\% in English.


\subsection{Coordination}
We further partition each dataset using a common coordination metric \cite{burghardt2023socio,luceri2023unmasking}, in which accounts are deemed coordinated if they post near-duplicate sequences of hashtags (which are strongly associated with near-duplicate messages). More specifically, for any pair of accounts, we label them as coordinated if (a) they both have posts containing three or more hashtags, and (b) the sequence of hashtags is exactly the same between both posts. While simple, this method has proven to be a very effective tool for detecting coordinated accounts \cite{burghardt2023socio}. This method is effective because even when coordinated accounts try to obfuscate their authenticity by re-writing a post across various posts, the sequence of hashtags is often the same. In addition, we create a network of accounts that are linked based on this indicator of coordination. Connected clusters often represent distinct sets of messages consistent with particular information operations.

Applying this method to the Israel-Hamas war dataset, we extract 4.2K coordinated accounts that posted 69K posts, of which 82.7\% were in English, 16.7\% in Arabic, 0.5\% in Hebrew, and a small proportion in other languages. For the 2022 French Election dataset, we extract 179 accounts that posted 27K posts, of which 91\% were in French and the rest were in English (a substantially smaller percentage that were in French compared to the data at large). To check the robustness of these results, we also used an alternative state-of-the-art coordination algorithm \cite{luceri2023unmasking}, in which we calculated the co-repost similarity, co-url similarity, rapid retweet similarity, text similarity, and the hashtags shared, and combined these into one network. Within this network, we calculated accounts as coordinated if their eigenvector centrality were > 0.002, following prior work that shows even this modest centrality metric strongly separates coordinated accounts from authentic users \cite{luceri2023unmasking}. This method finds 409 accounts that post 6.8K posts in the Israel-Hamas dataset and 1.8K accounts that post 9.2K posts in the French election dataset. In the former dataset, we find 99.6\% of coordinated account posts are in English, and the rest in Arabic, while in the latter dataset, 91\% of posts are in French and the rest in English. 

We share the coordinated networks in the repository, but, because X's terms of service do not allow us to share post text, and because we have removed all text IDs and account IDs, we do not share additional data. Because these data are public and anonymized, it was not necessary to obtain consent from accounts to extract these data. We show the frequency of posts over time for coordinated and authentic (non-coordinated) accounts in the Appendix (Figs.~\ref{fig:hamas-tweet-freq}).

\begin{figure*}[tbh!]
    \centering
    \includegraphics[width=1\linewidth]{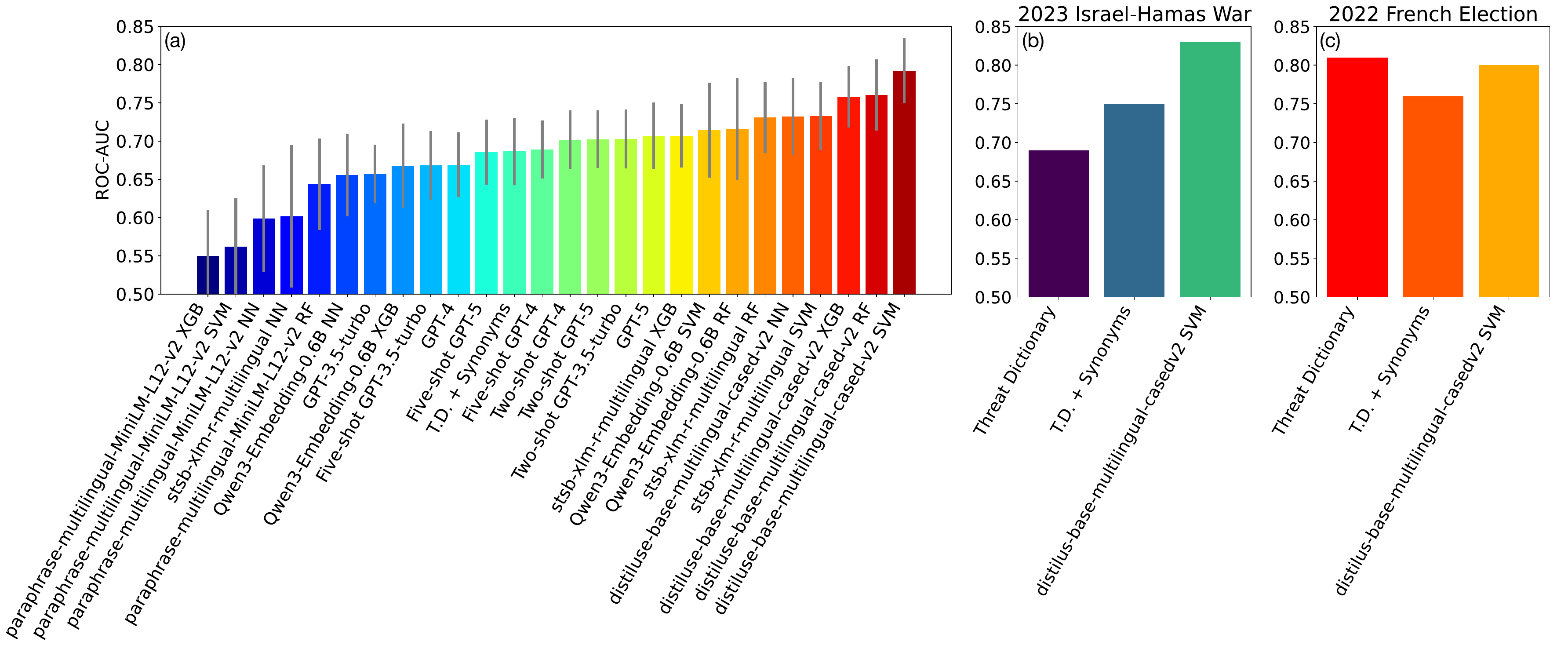}
    \caption{Performance of models on human-annotated X posts. (a) We show the ROC-AUC of XG-Boost (XGB) \cite{chen2016xgboost}, a neutral network (NN) \cite{tensorflow2015-whitepaper}, random forest (RF) \cite{ho1995random}, and a support vector machine (SVM) \cite{hearst1998support} trained on human-annotated posts for multi-lingual embedding models described in the Scientific Methods section. We also show the ROC-AUC of GPT-3.5, GPT-4, and GPT-5 with zero-shot, two-shot, and five-shot predictions. Finally, we also show performance of the Threat Dictionary plus synonyms (``T.D. + Synonyms'') at predicting threat text (if a post contained a word that was in this set of words, we labeled the data ``hazard''; otherwise it was not). Gray bars represent standard deviations across 50 evaluations. (b) Best model performance and baseline performance on posts from the 2023 Israel-Hamas war dataset. (c) Best model performance and baseline performance on posts from the 2022 French election dataset.
    }
    \label{fig:performance}
\end{figure*}

\section{Results}
\subsection{Hazards Model and Validation}
Performance of the hazards detection model on benchmark data is shown in Fig.~\ref{fig:performance}. Despite the simplicity of the XGBoost model, its performance is comparable to LLM models, with an area under the receiver operating characteristic curve (ROC-AUC) of 0.79 $\pm 0.04$ (Fig.~\ref{fig:performance}). Due to the variance in the performance metric (gray bars), it is uncertain whether the model outperforms all models, although its substantially higher throughput and low cost make it an obvious choice compared to GPT models when applied to millions of social media posts. This training dataset uses posts that contain words from the Threat Dictionary \cite{Choi2022}, making the dataset especially challenging, as, despite the posts containing threat words, only a subset include information about hazards. Notably, this implies that simple lexical models, like those based on the Threat Dictionary and synonyms, are a poorer tool, as the Threat Dictionary + Synonym baseline has an ROC-AUC of $0.68\pm0.04$. This demonstrates the need for an AI-based method that can go beyond dictionary baselines. We also compare our human annotations of posts in the Israel-Hamas and 2022 French election dataset, respectively (see details in the Research Methods section), where we translate all text to English to give dictionary-based methods a more even footing. The results indicate the model achieves strong generalizability, even across thematically distinct datasets and different years, and captures context better than dictionary approaches.

\begin{figure}[tbh!]
    \centering
    \includegraphics[width=0.93\columnwidth]{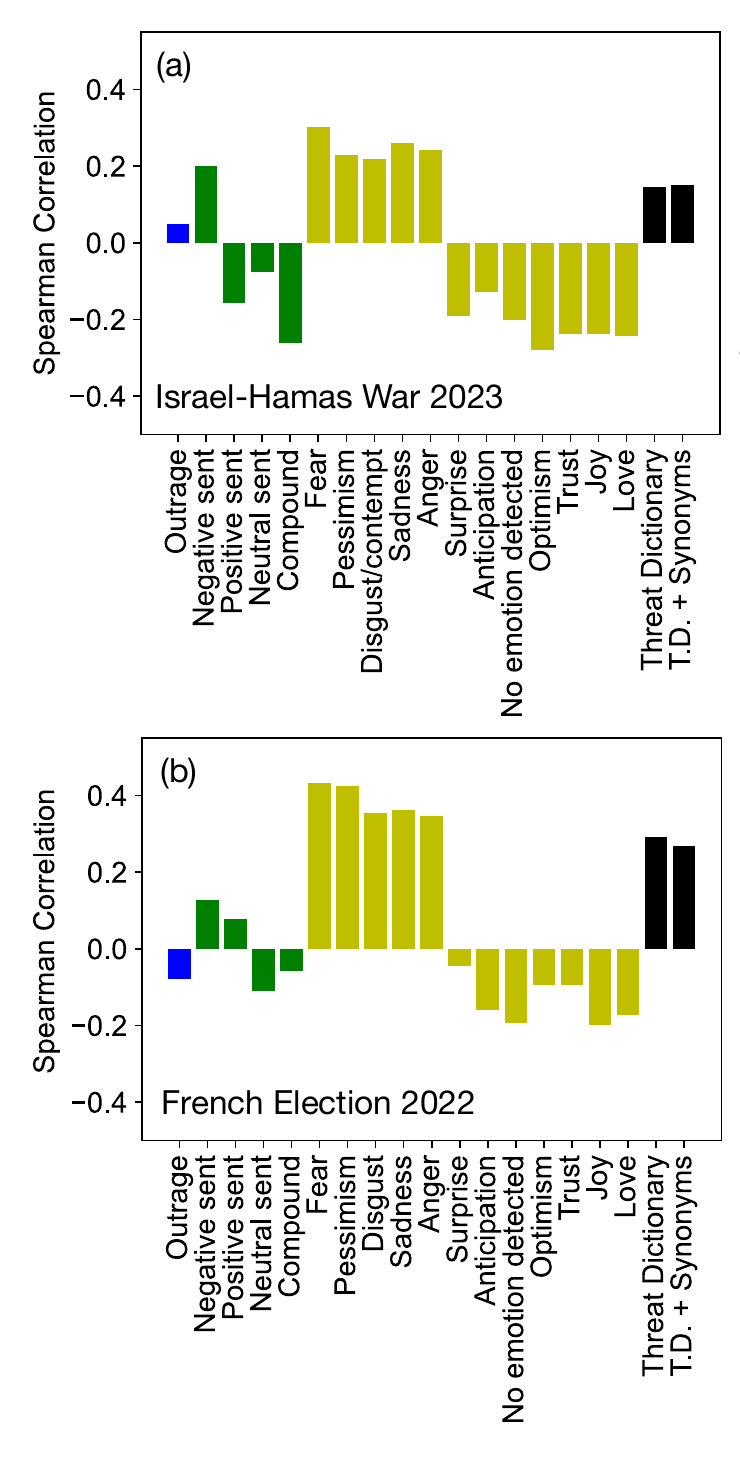}
    \caption{Understanding hazards. Spearman correlation between hazards moral outrage \cite{Brady2021}, sentiment \cite{hutto2014vader}, emotions \cite{Chochlakis2023}, Threat Dictionary \cite{lilienfeld2014threat} and threat synonyms for (a) the Israel-Hamas war, (b) the 2022 French election. All values are statistically significant (p-value $<0.05$).
    }
    \label{fig:correl}
\end{figure}

\subsection{Hazards in Real-World Data}
We show in Fig.~\ref{fig:correl} how the hazard confidences compare to other linguistic indicators of affect in our two X datasets on important geopolitical events. We then show the words most often seen in high- and low-hazard posts in each dataset, clarifying how hazards are typically described. Next, we show how hazards are discussed over time, especially within different sets of coordinated accounts that concern each geopolitical event. 

\subsubsection{Linguistic Analysis of Hazards}
As we detail in the Discussion section, our model can potentially be used to tackle a variety of questions in which the presence of hazard information is relevant. In order to demonstrate the model's utility as a research tool, below we employ the model to investigate coordinated accounts in social media.

Figure~\ref{fig:correl} shows correlations between hazard confidences and other text indicators of affect-relevant phenomena expressed in each post, including moral outrage~\cite{Brady2021}, sentiment~\cite{hutto2014vader}, emotion confidence \cite{Chochlakis2023}, and threat words \cite{Choi2022}. 
In all cases, the absolute value of Spearman correlations is below 0.5, suggesting that alternative indicators do not fully capture information about hazards in text. That said, the positive or negative direction of the correlations makes intuitive sense. For example, moral outrage is weakly positively correlated with hazards, perhaps because people share their moral outrage towards some hazards, such as those that harm the innocent. Similarly, negative sentiment, as well as most negative emotions and posts containing Threat Dictionary \cite{Choi2022} words, show positive correlations with hazards. This is consistent with the negative framing of the hazard posts (cf. example hazard annotations and predictions in the Appendix Table~\ref{tab:hazard_ex}).

We also find in Appendix Tables~\ref{tab:israel_hazards} \& ~\ref{tab:phase1b_hazards} that high-hazard post words relate to ``terror'', ``bien-\^etre'' (well-being), or ``corrupt'' while low-hazard post words include ``pond,'' ``demain'' (request), or "2019," which appear in posts that describe non-violent concepts, especially those that are not time-sensitive. 
While some hazard words are similar to threat words seen in \cite{Choi2022}, some hazard-related words cannot easily be captured with the Threat Dictionary, such as ``children," ``send,'' or ``American,'' as these are associated with the target of harms (harm of the innocent, sending aid, etc.). 

\begin{figure}[tbh!]
    \centering
    \includegraphics[width=1\columnwidth]{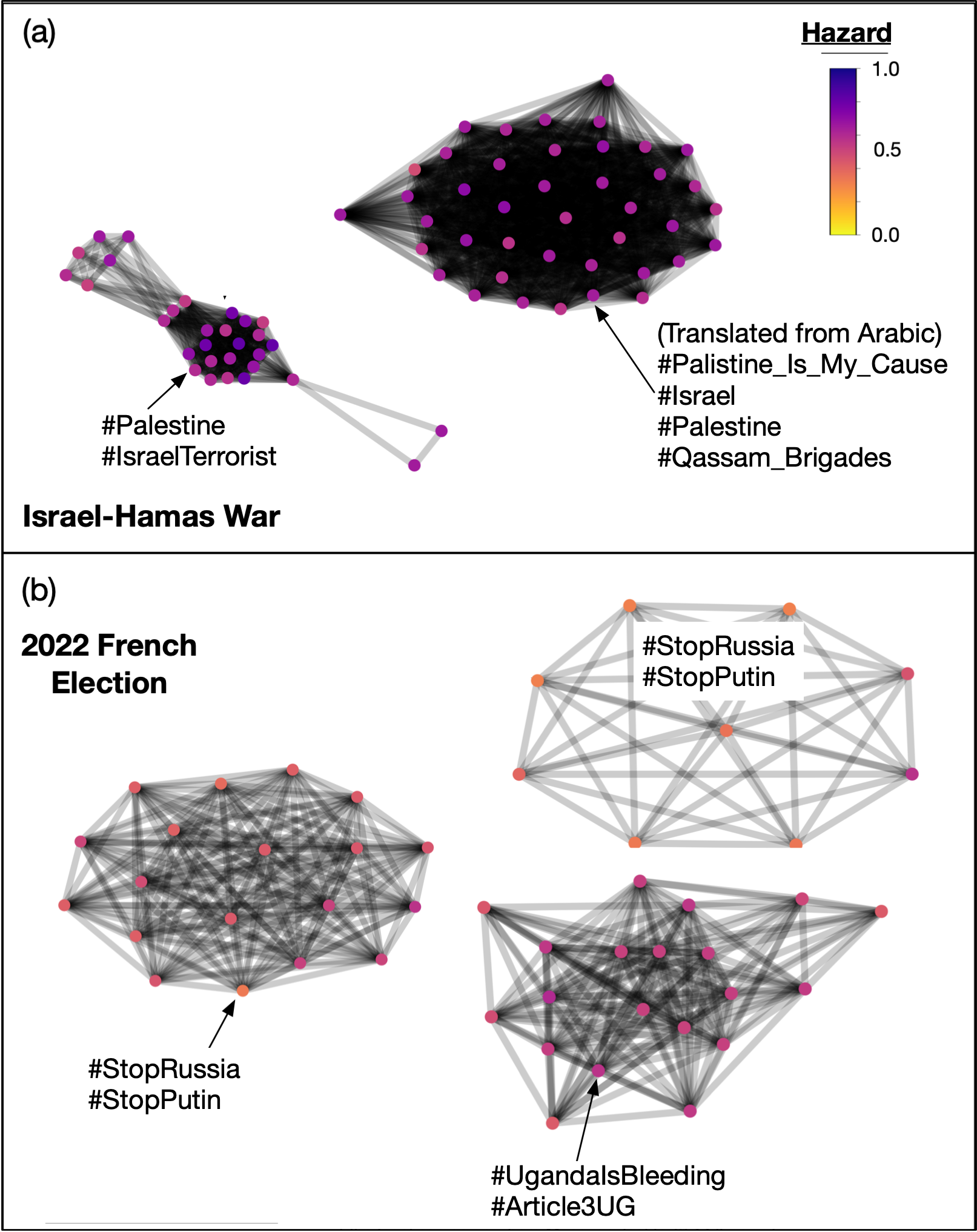}
    \caption{Hazards in representative large coordinated account clusters for (a) the Israel-Hamas war, (b) the 2022 French election. Mean hazard confidence per account is shown as a color from yellow to purple. Top hashtags and example posts are shown next to each cluster. 
 }
    \label{fig:coord-hazard}
\end{figure}

\subsubsection{Israel-Hamas War}
Next, we explore how hazards are used within coordinated influence campaigns, specifically how hazard content differs between authentic and inauthentic posts in the Israel-Hamas war dataset.

 Appendix Fig.~\ref{fig:haz_freq_coord-ih} shows that likely authentic accounts emphasize the association between hazards and children, civilians, and terror (reflecting hazards surrounding the October 7th Hamas attack on Israel, as well as later bombings of Gaza). In contrast, coordinated accounts do not appear to associate the terrorist attack with a hazard, but instead associate hazards with bombs and civilians (reflecting Israel's assault on Gaza). Analysis of individual posts confirms these observations. We show words associated with high- and low-hazard posts for coordinated and authentic accounts in Appendix Tables~\ref{tab:challenge_hazards_coord_hash},~\ref{tab:challenge_hazards_coord_merged},~\ref{tab:challenge_hazards_noncoord_hash} \&~\ref{tab:challenge_hazards_noncoord_merged}. We find that the hazard rate is not statistically significantly different overall between coordinated and authentic accounts (Mann-Whitney U test p-value $>0.1$), where both have a mean hazard confidence of 0.54 across all posts (including reposts), but this belies significant differences within clusters and across time. Moreover, with an alternative coordination metric \cite{luceri2023unmasking}, we find coordinated accounts have significantly higher mean model confidence value (0.55 vs 0.54, Mann-Whitney U test p-value $<0.0001$).

We look within coordinated accounts in Fig.~\ref{fig:coord-hazard}a, which shows examples of smaller coordinated account clusters. The links connect accounts that are deemed to be coordinated based on their text (see Methods). The top hashtags are shown next to each cluster. Both clusters in this figure show higher hazard confidence (0.55 and 0.63 for the left and right clusters, respectively, versus 0.54 for all non-coordinated clusters, Mann-Whitney U test p-value $<0.001$), apparently because of the discussion of hazards against civilians (e.g., ``\#IsraelTerrorist'' is the second-most popular hashtag in the cluster on the left). We show embeddings of posts from these clusters are shown in the Appendix Fig.~\ref{fig:hazard_embedding}a, where we see the posts have overlapping embedding distributions, suggesting the topic themes are similar \cite{grootendorst2022bertopic}, in agreement with the top hashtags observed.

The top hashtags for the largest cluster of our dataset (1.9K accounts, 42K posts, mean hazard confidence across all posts: 0.52) are '\#IsraeliNewNazism', (3.3K posts), '\#Gaza\_under\_attack' (2.3K posts), '\#Israel', (2.3K posts), and so on. Posts on October 7, 2023, the day of the Hamas attack, discount the Israeli lives lost with posts such as ``Palestinian rockets launched from Gaza hit the center of Israel’s capital city, Tel Aviv. Finally Israel Is Going To Be Finished Today Insha'Allah''. Posts long after the attack, meanwhile, discuss hazards toward Gazans, e.g., ``"My three children. I lost them all!"...'' (And interestingly, the same accounts promote protests in the U.S., with posts such as ``Enthusiastic demonstration of students and professors of the University of California, Berkeley in support of Palestine...''). In general, these results paint a picture of a selective discussion of hazards.

\subsubsection{2022 French Election}
As a second demonstration of the model's utility, we analyze social media related to the 2022 French election. Coordinated influence campaign accounts appear to hijack election discussions to request aid to Ukraine during the Russia-Ukraine war, as shown in Fig.~\ref{fig:coord-hazard}b, where we plot the largest clusters consisting of 48 out of 179 coordinated accounts. Account-level hazard rates do not differ substantially between coordinated and authentic accounts (0.35, Mann-Whitney U test p-value $>0.1$), although for an alternative coordination metric \cite{luceri2023unmasking}, we find coordinated accounts have significantly higher confidence (0.48 vs 0.35, Mann-Whitney U test p-value $<0.0001$). These findings match what we also see in the Israel-Hamas war dataset. Appendix Fig.~\ref{fig:haz_freq_coord-french} shows that likely authentic accounts emphasize the association between hazards and the Russia-Ukraine war (``russie'' and ``l'ukraine''), while coordinated accounts discuss nuclear (``nucl\'eaire'') or associate Russia with hazards ``russian'' vs low-hazard ``l'ukraine''. See also the words most, and least, associated with hazard posts in Appendix Tables~\ref{tab:phase1b_hazards_coord_hash},~\ref{tab:phase1b_hazards_coord_merged},~\ref{tab:phase1b_hazards_noncoord_hash} \& ~\ref{tab:phase1b_hazards_noncoord_merged}. To better understand these results, we analyze the individual clusters of coordinated accounts.

In Fig.~\ref{fig:coord-hazard}b, we see 3 major clusters. The second and third-largest of which (28 accounts, 361 posts, mean hazard confidence across all posts:  0.53 for the largest, and 0.45 for the smallest cluster, Mann-Whitney p-value $<0.001$, compared to non-coordinated posts) has the top hashtags \#StopRussia' (156 posts), \#StopPutin (154 posts). We see that the vast majority of these posts were aimed at a French audience, especially Emmanuel Macron, with posts such as ``@EmmanuelMacron Ban Russia from SWIFT!...'' (over 92\% of the posts in each cluster mention ``Macron'' or ``macron''). These posts are broadly requesting aid to stop Russia's invasion of Ukraine. In contrast, the largest cluster (20 accounts, 322 posts) hijacks terms associated with the French election to advocate for regime change in Uganda; it too has higher hazard model confidence values (0.46, Mann-Whitney U test p-value $<0.001$). We show embeddings of posts from these clusters are shown in the Appendix Fig.~\ref{fig:hazard_embedding}b, where we see the clusters related to Russia and Putin posts have overlapping embedding distributions, suggesting the topic themes are similar \cite{grootendorst2022bertopic}, in agreement with the top hashtags observed. The Uganda-themed cluster, meanwhile, shows a distinct distribution.

\subsection{Hazards Over Time}
The model also demonstrates several surprising trends in the way hazards vary over time.

\begin{figure}[tbh!]
    \centering
    \includegraphics[width=1\columnwidth]{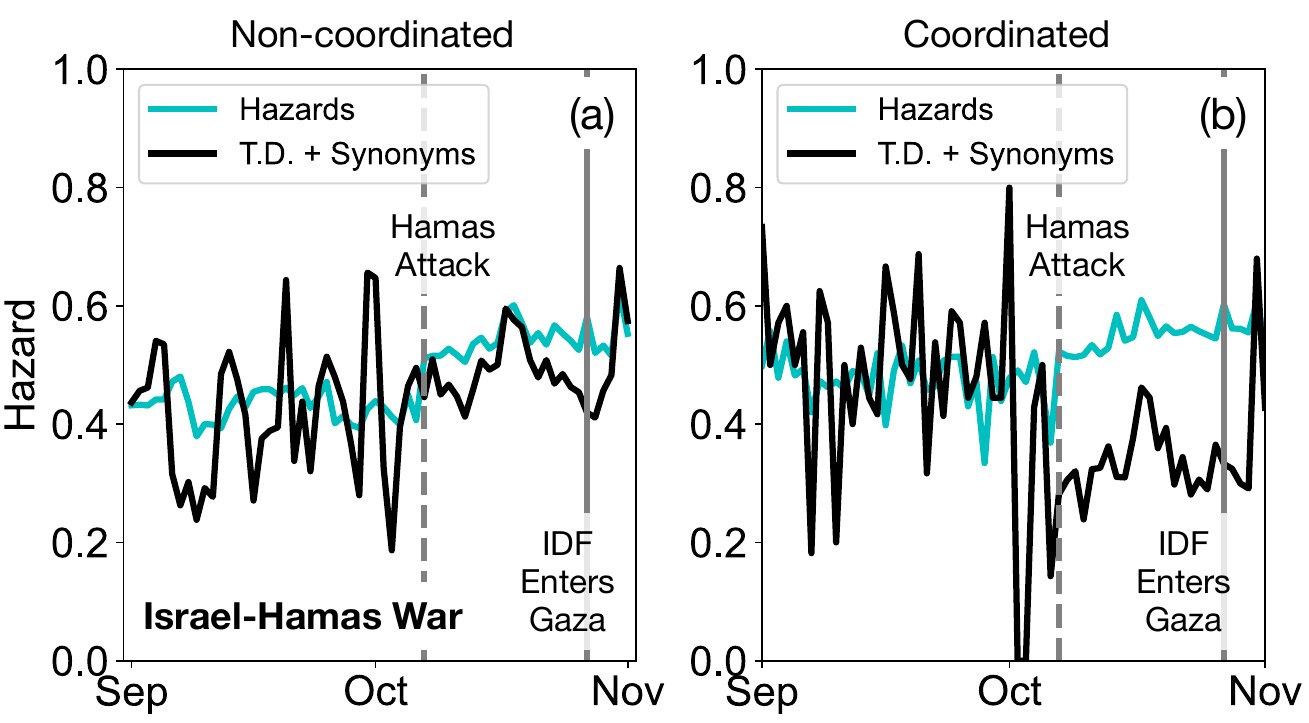}
    \caption{Hazards and threats over time for the Israel-Hamas war dataset. The plots show mean hazard confidences each day as well as the overall mean proportion of posts with at least one word from the Threat Dictionary \cite{Choi2022} + Synonyms baseline for (a) authentic accounts and (b) inauthentic coordinated accounts. Vertical lines correspond to the Hamas attack on October 7th, 2023, and the Israel Defense Forces (IDF) entering Gaza on October 27th.
 }
    \label{fig:hamas-hazard}
\end{figure}

We plot hazard confidence in posts over time in Fig.~\ref{fig:hamas-hazard} for the Israel-Hamas war dataset. Just after the October 7th attack, there is a sudden increase in mean hazard confidence among authentic accounts as they discuss the attack on Israeli civilians. There is no obvious increase in hazard discussions among coordinated accounts, although there is a decrease in threat words. However, there is a spike in the number of posts by coordinated accounts, shown in Appendix Fig.~\ref{fig:hamas-tweet-freq}. 
We also see only a minor change in the proportion of posts that contain threat words \cite{Choi2022} after the October 7th attack. Our hazard model can therefore constitute a distinct and potentially more accurate indicator of major hazard events compared to the previous state-of-the-art methods. In a separate analysis, coordinated account posts appear to show elation (promotion of Hamas, Appendix Fig.~\ref{fig:hamas-ment-freq}, and positive emotions, Appendix Figs.~\ref{fig:hamas-joy-freq} \&~\ref{fig:hamas-opt-freq}), thus the absence of a significant change in the frequency of hazards just after October 7th occurs despite the accounts mentioning the attacks, rather than because the accounts ignore the event.

Appendix Figs.~\ref{fig:phase1b-hazard_hash} \& ~\ref{fig:phase1b-hazard_merged}, meanwhile, shows hazards over time for both coordinated and authentic accounts in the 2022 French election dataset. We see notable dips during the election rounds, possibly because the vast majority of posts are positive, with posts related to ``vote''. 
We show in Appendix Fig.~\ref{fig:phase1b-tweet-freq}  that, with the exception of mid-May, we also see a sharp drop in coordinated account posts just after each election round. This could reflect the lower potential payoffs of information operations at these time points.

\section{Discussion}

Information concerning hazards is particularly potent, as it will frequently both garner more attention, and be more likely to be believed, than other types of information. We can therefore expect that information concerning hazards will be especially impactful in social media arenas -- and, correspondingly, that this type of information will be deployed as part of calculated, coordinated social media operations intended to influence public opinion and achieve strategic objectives. Although existing approaches are effective at identifying related phenomena, such as negative affective content or terms associated with threat, none suffice for the task of pinpointing hazard information. Addressing this gap, we have developed a transformer-based model to detect hazard information in social media posts. Our model outpaces simple word-based proxies, such as the Threat Dictionary \cite{Choi2022}, and matches sophisticated LLM-based approaches with far higher throughput. 

To demonstrate the model's utility, we applied it to large samples of X posts regarding, respectively, the 2023 Israel-Hamas war and the 2022 French election, 
both of which constitute critical recent geopolitical events, and each of which is distinct in the languages used, thus illustrating strengths of our model's ability to analyze multilingual social media data. Partitioning the data by accounts that are coordinated (and thus are likely part of information operations) and not coordinated (and thus are likely authentic accounts), we utilize our model to illuminate how hazard information is deployed in an information warfare environment. 

In the Israel-Hamas war dataset, we find that coordinated accounts focus on hazards facing Gazans over hazards facing Israelis, even after an attack on October 7 that killed almost 1,200 Israelis \cite{hamasattack}. In the 2022 French election dataset, we find that a substantial proportion of posts are pro-Ukraine and anti-Russian, including posts in both English and French that condemn the Bucha massacre and the Russian invasion. 

Overall, our analysis reveals that coordinated accounts on X often support a weaker group in a conflict, whether Hamas (opposing Israel) or Ukraine (opposing Russia). Moreover, these accounts often mention hazards impacting civilians, possibly in an attempt to evoke sympathy and enlist foreign support for their cause. 

\section{Limitations}

Although we are confident that identifying hazards in text has myriad critical applications, we caution that our model's performance could still be improved. Both the subjectivity of text annotations \cite{davani2022dealing} and a relatively small number of annotations reduce the accuracy of many text indicator models \cite{Chochlakis2023,Brady2021}, a problem likely also present in hazard detection. Finally, while we implemented several safeguards to maximize the validity of our crowdsourced annotations, we cannot guarantee that the annotations were from organic users. 
Furthermore, while the coordinated accounts are suspicious, and their behavior is suggestive of an information operation, we cannot guarantee the true intention of coordinated users nor whether they are inauthentic users. There is an inevitable gray area between users who happen to post geopolitical content, even simultaneously, and users who try to influence geopolitical events. 

These limitations indicate the need to iteratively improve upon our hazard detection model. These improvements can include more multilingual human-annotated data, possibly augmented with annotations by LLMs, to increase the generalizability of a model. We can also develop a multi-modal (text, images, and video) model to detect hazard information, to capture hazards in, for example, visual memes or to illuminate how traditional television media portray hazards, or use hazard information for their own editorial goals. 
Finally, we need to find ways to better distinguish information operations of, e.g., state actors, from other types of users so that we can better understand the intentions of these actors, especially in how they utilize hazards in text. Analysis of these data can include analyzing text clusters akin to BERTopic \cite{grootendorst2022bertopic}. This could uncover patterns in coordination beyond supervised labels. 

\section{Conclusion}
Humans have a tendency to respond to, and share, information about hazardous events. 
In this work, we develop AI tools with which it is possible to observe the implications or manifestations of these tendencies at scale in online social media. Our findings are as follows. 
   (1) We develop an openly shared tool to accurately detect information concerning hazards at scale. 
   (2) We use this tool to reveal how this information associates with, but goes beyond, negative emotions and sentiments, as well as threat words. 
    (3) We apply this tool to X posts about two geopolitical events, finding that information operations supporting weaker parties in conflicts often mention harms directed at civilians, potentially in an attempt to evoke sympathy and attract support for their cause. 

While these findings demonstrate the utility of detecting information concerning hazards, we view our work as merely the first step, as we envision myriad applications of our model and subsequent improvements thereon, both in CS research, and in a wide variety of related fields. Given that rapidly diffusing and highly motivating hazard information has the potential to enhance large-scale commitment to collective goals, it is vital that investigators shed light on the ways that such information is deployed in online contexts, as the same features of human psychology that present an opportunity for promoting broad cooperation toward the common good also constitute a vulnerability that can be exploited by those seeking to manipulate audiences for their own purposes. 

\section*{Ethical Statement}
This work was approved by an IRB. To improve account privacy, we removed all personally identifiable information from the data, such as post IDs and account IDs, prior to our analysis. Although the hazard model performs well at scale, it is still possible for the model to make mistakes. Therefore, care must be taken when interpreting the model output, including when applying it to detect whether individual accounts are sharing information concerning hazards. The confidence of the model does not guarantee that individual accounts or posts are sharing hazard content, especially given that sarcasm and in-group language constitute a challenge for current AI models. However, much like a sentiment- or emotion-detection model, so long as researchers are aware of these limitations, the harm to society is minimal; therefore, we share this model widely via the link presented at the top of the paper. We believe that the utility to society of the hazard detection model far outweighs any harmful societal impact.

\section*{Acknowledgments}
This work was supported in part by NSF under award 2331722 and DARPA under contract HR001121C0168. We would also like to thank Daniel Penn, Chloe Keshishian, and Anika Scott for their help annotating text.


\begin{thebibliography}{78}
\providecommand{\natexlab}[1]{#1}

\bibitem[{Abadi et~al.(2015)Abadi, Agarwal, Barham, Brevdo, Chen, Citro,
  Corrado, Davis, Dean, Devin et~al.}]{tensorflow2015-whitepaper}
Abadi, M.; Agarwal, A.; Barham, P.; Brevdo, E.; Chen, Z.; Citro, C.; Corrado,
  G.~S.; Davis, A.; Dean, J.; Devin, M.; et~al. 2015.
\newblock {TensorFlow}: Large-Scale Machine Learning on Heterogeneous Systems.
\newblock Software available from tensorflow.org.

\bibitem[{Acheampong, Nunoo-Mensah, and Chen(2021)}]{acheampong2021transformer}
Acheampong, F.~A.; Nunoo-Mensah, H.; and Chen, W. 2021.
\newblock Transformer models for text-based emotion detection: a review of
  BERT-based approaches.
\newblock \emph{Artificial Intelligence Review}, 54(8): 5789--5829.

\bibitem[{Achiam et~al.(2023)Achiam, Adler, Agarwal, Ahmad, Akkaya, Aleman,
  Almeida, Altenschmidt, Altman, Anadkat et~al.}]{achiam2023gpt}
Achiam, J.; Adler, S.; Agarwal, S.; Ahmad, L.; Akkaya, I.; Aleman, F.~L.;
  Almeida, D.; Altenschmidt, J.; Altman, S.; Anadkat, S.; et~al. 2023.
\newblock GPT-4 technical report.
\newblock \emph{arXiv preprint arXiv:2303.08774}.

\bibitem[{Alhuzali and Ananiadou(2021)}]{alhuzali2021spanemo}
Alhuzali, H.; and Ananiadou, S. 2021.
\newblock SpanEmo: Casting multi-label emotion classification as
  span-prediction.
\newblock \emph{arXiv preprint arXiv:2101.10038}.

\bibitem[{Alshehri, Abdul-Mageed et~al.(2020)}]{alshehri2020understanding}
Alshehri, A.; Abdul-Mageed, M.; et~al. 2020.
\newblock Understanding and detecting dangerous speech in social media.
\newblock In \emph{Proceedings of the 4th Workshop on Open-Source Arabic
  Corpora and Processing Tools, with a Shared Task on Offensive Language
  Detection}, 40--47.

\bibitem[{authors(2016)}]{gpyopt2016}
authors, T.~G. 2016.
\newblock GPyOpt: A Bayesian Optimization framework in Python.
\newblock \url{http://github.com/SheffieldML/GPyOpt}.

\bibitem[{Badawy, Ferrara, and Lerman(2018)}]{badawy2018analyzing}
Badawy, A.; Ferrara, E.; and Lerman, K. 2018.
\newblock Analyzing the digital traces of political manipulation: The 2016
  Russian interference Twitter campaign.
\newblock In \emph{ASONAM}, 258--265. IEEE.

\bibitem[{Bago et~al.(2021)Bago, Rosenzweig, Berinsky, and
  Rand}]{bago2021emotion}
Bago, B.; Rosenzweig, L.; Berinsky, A.~J.; and Rand, D.~W. 2021.
\newblock Emotion may predict susceptibility to fake news but emotion
  regulation does not help.
\newblock \emph{IAST working paper}.

\bibitem[{Baumeister et~al.(2001)Baumeister, Bratslavsky, Finkenauer, and
  Vohs}]{baumeister2001bad}
Baumeister, R.~F.; Bratslavsky, E.; Finkenauer, C.; and Vohs, K.~D. 2001.
\newblock Bad is stronger than good.
\newblock \emph{Review of general psychology}, 5(4): 323--370.

\bibitem[{Bebbington et~al.(2017)Bebbington, MacLeod, Ellison, and
  Fay}]{bebbington2017sky}
Bebbington, K.; MacLeod, C.; Ellison, T.~M.; and Fay, N. 2017.
\newblock The sky is falling: evidence of a negativity bias in the social
  transmission of information.
\newblock \emph{Evolution and Human Behavior}, 38(1): 92--101.

\bibitem[{Bird and Loper(2004)}]{bird-loper-2004-nltk}
Bird, S.; and Loper, E. 2004.
\newblock {NLTK}: The Natural Language Toolkit.
\newblock In \emph{Proceedings of the {ACL} Interactive Poster and
  Demonstration Sessions}, 214--217. Barcelona, Spain: Association for
  Computational Linguistics.

\bibitem[{Blaine and Boyer(2018)}]{blaine2018origins}
Blaine, T.; and Boyer, P. 2018.
\newblock Origins of sinister rumors: A preference for threat-related material
  in the supply and demand of information.
\newblock \emph{Evolution and Human Behavior}, 39(1): 67--75.

\bibitem[{Blinken(2024)}]{hamasattack}
Blinken, A.~J. 2024.
\newblock Anniversary of October 7th Attack.
\newblock \emph{Press Statement},
  \url{https://www.state.gov/anniversary--of--october--7th--attack/}.

\bibitem[{Brady et~al.(2021)Brady, McLoughlin, Doan, and Crockett}]{Brady2021}
Brady, W.~J.; McLoughlin, K.; Doan, T.~N.; and Crockett, M.~J. 2021.
\newblock How social learning amplifies moral outrage expression in online
  social networks.
\newblock \emph{Science Advances}, 7(33): eabe5641.

\bibitem[{Brady et~al.(2017)Brady, Wills, Jost, Tucker, and
  Van~Bavel}]{brady2017emotion}
Brady, W.~J.; Wills, J.~A.; Jost, J.~T.; Tucker, J.~A.; and Van~Bavel, J.~J.
  2017.
\newblock Emotion shapes the diffusion of moralized content in social networks.
\newblock \emph{PNAS}, 114(28): 7313--7318.

\bibitem[{Brueckl and Heuer(2022)}]{irrna}
Brueckl, M.; and Heuer, F. 2022.
\newblock \emph{irrNA: Coefficients of Interrater Reliability – Generalized
  for Randomly Incomplete Datasets}.
\newblock R package version 0.2.3.

\bibitem[{Burghardt, Hogg, and Lerman(2018)}]{burghardt2018quantifying}
Burghardt, K.; Hogg, T.; and Lerman, K. 2018.
\newblock Quantifying the impact of cognitive biases in question-answering
  systems.
\newblock In \emph{{CSCW}}, volume~12, 568--571.

\bibitem[{Burghardt et~al.(2023)Burghardt, Rao, Guo, He, Chochlakis,
  Sabyasachee, Rojecki, Narayanan, and Lerman}]{burghardt2023socio}
Burghardt, K.; Rao, A.; Guo, S.; He, Z.; Chochlakis, G.; Sabyasachee, B.;
  Rojecki, A.; Narayanan, S.; and Lerman, K. 2023.
\newblock Socio-Linguistic Characteristics of Coordinated Inauthentic Accounts.
\newblock \emph{arXiv preprint arXiv:2305.11867}.

\bibitem[{Cao et~al.(2025)Cao, Hong, Li, Ying, Ma, Liang, Liu, Yao, Wang, Huang
  et~al.}]{cao2025toward}
Cao, Y.; Hong, S.; Li, X.; Ying, J.; Ma, Y.; Liang, H.; Liu, Y.; Yao, Z.; Wang,
  X.; Huang, D.; et~al. 2025.
\newblock Toward generalizable evaluation in the llm era: A survey beyond
  benchmarks.
\newblock \emph{arXiv preprint arXiv:2504.18838}.

\bibitem[{Chen and Guestrin(2016)}]{chen2016xgboost}
Chen, T.; and Guestrin, C. 2016.
\newblock Xgboost: A scalable tree boosting system.
\newblock In \emph{SIGKDD}, 785--794.

\bibitem[{Chochlakis et~al.(2023)Chochlakis, Mahajan, Baruah, Burghardt,
  Lerman, and Narayanan}]{Chochlakis2023}
Chochlakis, G.; Mahajan, G.; Baruah, S.; Burghardt, K.; Lerman, K.; and
  Narayanan, S. 2023.
\newblock Leveraging Label Correlations in a Multi-Label Setting: a Case Study
  in Emotion.
\newblock In \emph{ICASSP}, 1--5.

\bibitem[{Choi et~al.(2022)Choi, Shrestha, Pan, and Gelfand}]{Choi2022}
Choi, V.~K.; Shrestha, S.; Pan, X.; and Gelfand, M.~J. 2022.
\newblock When danger strikes: A linguistic tool for tracking America's
  collective response to threats.
\newblock \emph{PNAS}, 119(4): e2113891119.

\bibitem[{Davani, D{\'\i}az, and Prabhakaran(2022)}]{davani2022dealing}
Davani, A.~M.; D{\'\i}az, M.; and Prabhakaran, V. 2022.
\newblock Dealing with disagreements: Looking beyond the majority vote in
  subjective annotations.
\newblock \emph{Transactions of the Association for Computational Linguistics},
  10: 92--110.

\bibitem[{Dettmers et~al.(2023)Dettmers, Pagnoni, Holtzman, and
  Zettlemoyer}]{dettmers2023qlora}
Dettmers, T.; Pagnoni, A.; Holtzman, A.; and Zettlemoyer, L. 2023.
\newblock Qlora: Efficient finetuning of quantized llms.
\newblock \emph{Advances in neural information processing systems}, 36:
  10088--10115.

\bibitem[{Devlin et~al.(2018)Devlin, Chang, Lee, and
  Toutanova}]{devlin2018bert}
Devlin, J.; Chang, M.-W.; Lee, K.; and Toutanova, K. 2018.
\newblock Bert: Pre-training of deep bidirectional transformers for language
  understanding.
\newblock \emph{arXiv preprint arXiv:1810.04805}.

\bibitem[{Dey, Luceri, and Ferrara(2024)}]{dey2024coordinated}
Dey, P.; Luceri, L.; and Ferrara, E. 2024.
\newblock Coordinated activity modulates the behavior and emotions of organic
  users: A case study on tweets about the Gaza conflict.
\newblock In \emph{Companion Proceedings of the ACM Web Conference 2024},
  682--685.

\bibitem[{Dey and Salem(2017)}]{dey2017gate}
Dey, R.; and Salem, F.~M. 2017.
\newblock Gate-variants of gated recurrent unit (GRU) neural networks.
\newblock In \emph{2017 IEEE 60th international midwest symposium on circuits
  and systems (MWSCAS)}, 1597--1600. IEEE.

\bibitem[{du~Boisberranger(2024{\natexlab{a}})}]{RF}
du~Boisberranger, J. 2024{\natexlab{a}}.
\newblock RandomForestClassifier.
\newblock
  \url{https://scikit-learn.org/stable/modules/generated/sklearn.ensemble.RandomForestClassifier.html}.

\bibitem[{du~Boisberranger(2024{\natexlab{b}})}]{SVM}
du~Boisberranger, J. 2024{\natexlab{b}}.
\newblock SVC,
  \url{https://scikit-learn.org/dev/modules/generated/sklearn.svm.SVC.html}.

\bibitem[{Duppada, Jain, and Hiray(2018)}]{duppada2018seernet}
Duppada, V.; Jain, R.; and Hiray, S. 2018.
\newblock Seernet at semeval-2018 task 1: Domain adaptation for affect in
  tweets.
\newblock \emph{arXiv preprint arXiv:1804.06137}.

\bibitem[{Ecker et~al.(2022)Ecker, Lewandowsky, Cook, Schmid, Fazio, Brashier,
  Kendeou, Vraga, and Amazeen}]{ecker2022psychological}
Ecker, U.~K.; Lewandowsky, S.; Cook, J.; Schmid, P.; Fazio, L.~K.; Brashier,
  N.; Kendeou, P.; Vraga, E.~K.; and Amazeen, M.~A. 2022.
\newblock The psychological drivers of misinformation belief and its resistance
  to correction.
\newblock \emph{Nature Reviews Psychology}, 1(1): 13--29.

\bibitem[{Fay et~al.(2021)Fay, Walker, Kashima, and Perfors}]{fay2021socially}
Fay, N.; Walker, B.; Kashima, Y.; and Perfors, A. 2021.
\newblock Socially situated transmission: The bias to transmit negative
  information is moderated by the social context.
\newblock \emph{Cognitive Science}, 45(9): e13033.

\bibitem[{Felbo et~al.(2017)Felbo, Mislove, S{\o}gaard, Rahwan, and
  Lehmann}]{felbo2017using}
Felbo, B.; Mislove, A.; S{\o}gaard, A.; Rahwan, I.; and Lehmann, S. 2017.
\newblock Using millions of emoji occurrences to learn any-domain
  representations for detecting sentiment, emotion and sarcasm.
\newblock \emph{arXiv preprint arXiv:1708.00524}.

\bibitem[{Ferrara and Yang(2015)}]{ferrara2015quantifying}
Ferrara, E.; and Yang, Z. 2015.
\newblock Quantifying the effect of sentiment on information diffusion in
  social media.
\newblock \emph{PeerJ Computer Science}, 1: e26.

\bibitem[{Fessler(2019)}]{fessler2019believing}
Fessler, D. 2019.
\newblock Believing chicken little: Evolutionary perspectives on credulity and
  danger.
\newblock \emph{DRUMS: Distortions, rumours, untruths, misinformation \&
  smears}, 17--36.

\bibitem[{Fessler, Pisor, and Holbrook(2017)}]{Fessler2015}
Fessler, D. M.~T.; Pisor, A.~C.; and Holbrook, C. 2017.
\newblock Political Orientation Predicts Credulity Regarding Putative Hazards.
\newblock \emph{Psychological Science}, 28(5): 651--660.
\newblock PMID: 28362568.

\bibitem[{Fessler, Pisor, and Navarrete(2014)}]{Fessler2014}
Fessler, D. M.~T.; Pisor, A.~C.; and Navarrete, C.~D. 2014.
\newblock Negatively-Biased Credulity and the Cultural Evolution of Beliefs.
\newblock \emph{PLOS ONE}, 9(4): 1--8.

\bibitem[{{FORCE11}(2020)}]{fair}
{FORCE11}. 2020.
\newblock The FAIR Data principles.
\newblock \url{https://force11.org/info/the-fair-data-principles/}.

\bibitem[{Forgas(2019)}]{forgas2019role}
Forgas, J.~P. 2019.
\newblock On the role of affect in gullibility: Can positive mood increase, and
  negative mood reduce credulity?
\newblock \emph{The social psychology of gullibility}, 179--197.

\bibitem[{Gebru et~al.(2021)Gebru, Morgenstern, Vecchione, Vaughan, Wallach,
  III, and Crawford}]{gebru2021datasheets}
Gebru, T.; Morgenstern, J.; Vecchione, B.; Vaughan, J.~W.; Wallach, H.; III,
  H.~D.; and Crawford, K. 2021.
\newblock Datasheets for datasets.
\newblock \emph{Commun. ACM}, 64(12): 86–92.

\bibitem[{Gerard, Weninger, and Lerman(2025)}]{gerard2025fear}
Gerard, P.; Weninger, T.; and Lerman, K. 2025.
\newblock Fear and Loathing on the Frontline: Decoding the Language of Othering
  by Russia-Ukraine War Bloggers.
\newblock In \emph{Proceedings of the International AAAI Conference on Web and
  Social Media}, volume~19, 615--635.

\bibitem[{Grootendorst(2022)}]{grootendorst2022bertopic}
Grootendorst, M. 2022.
\newblock BERTopic: Neural topic modeling with a class-based TF-IDF procedure.
\newblock \emph{arXiv preprint arXiv:2203.05794}.

\bibitem[{Hama, Otsuka, and Ishii(2024)}]{Hama2024}
Hama, K.; Otsuka, A.; and Ishii, R. 2024.
\newblock Emotion Recognition in Conversation with Multi-step Prompting Using
  Large Language Model.
\newblock In Coman, A.; and Vasilache, S., eds., \emph{Social Computing and
  Social Media}, 338--346. Cham: Springer Nature Switzerland.
\newblock ISBN 978-3-031-61281-7.

\bibitem[{Head et~al.(2024)Head, MechCoder, Louppe, Shcherbatyi, fcharras,
  Vinícius, cmmalone, Schröder, nel215, Campos, Young, Cereda, Fan,
  Schwabedal, Hvass-Labs, Pak, SoManyUsernamesTaken, Callaway, Estève, Besson,
  Landwehr, Komarov, Cherti, Shi, Pfannschmidt, Linzberger, Cauet, Gut,
  Mueller, and Fabisch}]{skopt}
Head, T.; MechCoder; Louppe, G.; Shcherbatyi, I.; fcharras; Vinícius, Z.;
  cmmalone; Schröder, C.; nel215; Campos, N.; Young, T.; Cereda, S.; Fan, T.;
  Schwabedal, J.; Hvass-Labs; Pak, M.; SoManyUsernamesTaken; Callaway, F.;
  Estève, L.; Besson, L.; Landwehr, P.~M.; Komarov, P.; Cherti, M.; Shi,
  K.~K.; Pfannschmidt, K.; Linzberger, F.; Cauet, C.; Gut, A.; Mueller, A.; and
  Fabisch, A. 2024.
\newblock scikit-optimize: Sequential model-based optimization in Python.

\bibitem[{Hearst et~al.(1998)Hearst, Dumais, Osuna, Platt, and
  Scholkopf}]{hearst1998support}
Hearst, M.~A.; Dumais, S.~T.; Osuna, E.; Platt, J.; and Scholkopf, B. 1998.
\newblock Support vector machines.
\newblock \emph{IEEE Intelligent Systems and their applications}, 13(4):
  18--28.

\bibitem[{Ho(1995)}]{ho1995random}
Ho, T.~K. 1995.
\newblock Random decision forests.
\newblock In \emph{Proceedings of 3rd international conference on document
  analysis and recognition}, volume~1, 278--282. IEEE.

\bibitem[{Hochreiter and Schmidhuber(1997)}]{hochreiter1997long}
Hochreiter, S.; and Schmidhuber, J. 1997.
\newblock Long short-term memory.
\newblock \emph{Neural computation}, 9(8): 1735--1780.

\bibitem[{Hong et~al.(2025)Hong, Gong, Sethu, and Dang}]{Hong2025}
Hong, X.; Gong, Y.; Sethu, V.; and Dang, T. 2025.
\newblock AER-LLM: Ambiguity-aware Emotion Recognition Leveraging Large
  Language Models.
\newblock In \emph{ICASSP 2025 - 2025 IEEE International Conference on
  Acoustics, Speech and Signal Processing (ICASSP)}, 1--5.

\bibitem[{Hu et~al.(2022)Hu, Shen, Wallis, Allen-Zhu, Li, Wang, Wang, Chen
  et~al.}]{hu2022lora}
Hu, E.~J.; Shen, Y.; Wallis, P.; Allen-Zhu, Z.; Li, Y.; Wang, S.; Wang, L.;
  Chen, W.; et~al. 2022.
\newblock Lora: Low-rank adaptation of large language models.
\newblock \emph{ICLR}, 1(2): 3.

\bibitem[{Hutto and Gilbert(2014)}]{hutto2014vader}
Hutto, C.; and Gilbert, E. 2014.
\newblock Vader: A parsimonious rule-based model for sentiment analysis of
  social media text.
\newblock In \emph{{CSCW}}, volume~8, 216--225.

\bibitem[{Jamil, Pais, and Cordeiro(2022)}]{jamil2022detection}
Jamil, M.~L.; Pais, S.; and Cordeiro, J. 2022.
\newblock Detection of dangerous events on social media: a critical review.
\newblock \emph{Social Network Analysis and Mining}, 12(1): 154.

\bibitem[{Kartik, Ottaviani, and Squintani(2007)}]{Kartik2007}
Kartik, N.; Ottaviani, M.; and Squintani, F. 2007.
\newblock Credulity, lies, and costly talk.
\newblock \emph{Journal of Economic Theory}, 134(1): 93--116.

\bibitem[{Lilienfeld and Latzman(2014)}]{lilienfeld2014threat}
Lilienfeld, S.~O.; and Latzman, R.~D. 2014.
\newblock Threat bias, not negativity bias, underpins differences in political
  ideology.
\newblock \emph{Behavioral \& Brain Sciences}, 37(3): 318.

\bibitem[{Little(2017)}]{Little2017}
Little, A.~T. 2017.
\newblock Propaganda and credulity.
\newblock \emph{GEB}, 102: 224--232.

\bibitem[{Liu et~al.(2021)Liu, Ji, Fu, Tam, Du, Yang, and Tang}]{liu2021p}
Liu, X.; Ji, K.; Fu, Y.; Tam, W.~L.; Du, Z.; Yang, Z.; and Tang, J. 2021.
\newblock P-tuning v2: Prompt tuning can be comparable to fine-tuning
  universally across scales and tasks.
\newblock \emph{arXiv preprint arXiv:2110.07602}.

\bibitem[{Luceri et~al.(2024)Luceri, Pant{\`e}, Burghardt, and
  Ferrara}]{luceri2023unmasking}
Luceri, L.; Pant{\`e}, V.; Burghardt, K.; and Ferrara, E. 2024.
\newblock Unmasking the web of deceit: Uncovering coordinated activity to
  expose information operations on Twitter.
\newblock In \emph{Proceedings of the ACM Web Conference 2024}, 2530--2541.

\bibitem[{Martel, Pennycook, and Rand(2020)}]{Martel2020}
Martel, C.; Pennycook, G.; and Rand, D.~G. 2020.
\newblock Reliance on emotion promotes belief in fake news.
\newblock \emph{CR:PI}, 5(1): 47.

\bibitem[{Mazza et~al.(2019)Mazza, Cresci, Avvenuti, Quattrociocchi, and
  Tesconi}]{mazza2019rtbust}
Mazza, M.; Cresci, S.; Avvenuti, M.; Quattrociocchi, W.; and Tesconi, M. 2019.
\newblock Rtbust: Exploiting temporal patterns for botnet detection on twitter.
\newblock In \emph{Proceedings of the 10th ACM conference on web science},
  183--192.

\bibitem[{McInnes, Healy, and Melville(2018)}]{mcinnes2018umap}
McInnes, L.; Healy, J.; and Melville, J. 2018.
\newblock Umap: Uniform manifold approximation and projection for dimension
  reduction.
\newblock \emph{arXiv preprint arXiv:1802.03426}.

\bibitem[{Mendes and Martins(2023)}]{mendes2023quantifying}
Mendes, G.~A.; and Martins, B. 2023.
\newblock Quantifying valence and arousal in text with multilingual pre-trained
  transformers.
\newblock In \emph{ECIR}, 84--100. Springer.

\bibitem[{Mohammad et~al.(2018)Mohammad, Bravo-Marquez, Salameh, and
  Kiritchenko}]{mohammad2018semeval}
Mohammad, S.; Bravo-Marquez, F.; Salameh, M.; and Kiritchenko, S. 2018.
\newblock {SemEval}-2018 task 1: Affect in tweets.
\newblock In \emph{SemEval}, 1--17.

\bibitem[{{\"O}hman and Mineka(2001)}]{ohman2001fears}
{\"O}hman, A.; and Mineka, S. 2001.
\newblock Fears, phobias, and preparedness: toward an evolved module of fear
  and fear learning.
\newblock \emph{Psychological review}, 108(3): 483.

\bibitem[{Pedregosa et~al.(2011)Pedregosa, Varoquaux, Gramfort, Michel,
  Thirion, Grisel, Blondel, Prettenhofer, Weiss, Dubourg, Vanderplas, Passos,
  Cournapeau, Brucher, Perrot, and Duchesnay}]{scikit-learn}
Pedregosa, F.; Varoquaux, G.; Gramfort, A.; Michel, V.; Thirion, B.; Grisel,
  O.; Blondel, M.; Prettenhofer, P.; Weiss, R.; Dubourg, V.; Vanderplas, J.;
  Passos, A.; Cournapeau, D.; Brucher, M.; Perrot, M.; and Duchesnay, E. 2011.
\newblock Scikit-learn: Machine Learning in {P}ython.
\newblock \emph{JMRL}, 12: 2825--2830.

\bibitem[{Peng et~al.(2024)Peng, Zhang, Pang, Han, Zhao, Chen, and
  Schuller}]{peng2024customising}
Peng, L.; Zhang, Z.; Pang, T.; Han, J.; Zhao, H.; Chen, H.; and Schuller, B.~W.
  2024.
\newblock Customising General Large Language Models for Specialised Emotion
  Recognition Tasks.
\newblock In \emph{ICASSP}, 11326--11330. IEEE.

\bibitem[{Pennebaker, Francis, and Booth(2001)}]{pennebaker2001linguistic}
Pennebaker, J.~W.; Francis, M.~E.; and Booth, R.~J. 2001.
\newblock Linguistic inquiry and word count: LIWC 2001.
\newblock \emph{Mahway: Lawrence Erlbaum Associates}, 71(2001): 2001.

\bibitem[{Phillips et~al.(2025)Phillips, Wang, Carley, Rand, and
  Pennycook}]{phillips2024emotional}
Phillips, S.~C.; Wang, S. Y.~N.; Carley, K.~M.; Rand, D.~G.; and Pennycook, G.
  2025.
\newblock Emotional language reduces belief in false claims.
\newblock \emph{Judgment and Decision Making}, 20: e43.

\bibitem[{Reimers and Gurevych(2019)}]{reimers2019sentence}
Reimers, N.; and Gurevych, I. 2019.
\newblock Sentence-bert: Sentence embeddings using siamese bert-networks.
\newblock \emph{arXiv preprint arXiv:1908.10084}.

\bibitem[{Rozin and Royzman(2001)}]{rozin2001negativity}
Rozin, P.; and Royzman, E.~B. 2001.
\newblock Negativity bias, negativity dominance, and contagion.
\newblock \emph{PSPR}, 5(4): 296--320.

\bibitem[{Sachdeva et~al.(2022)Sachdeva, Barreto, Bacon, Sahn, von Vacano, and
  Kennedy}]{sachdeva-etal-2022-measuring}
Sachdeva, P.; Barreto, R.; Bacon, G.; Sahn, A.; von Vacano, C.; and Kennedy, C.
  2022.
\newblock The Measuring Hate Speech Corpus: Leveraging Rasch Measurement Theory
  for Data Perspectivism.
\newblock In Abercrombie, G.; Basile, V.; Tonelli, S.; Rieser, V.; and Uma, A.,
  eds., \emph{NLPerspectives}, 83--94. Marseille, France: ELRA.

\bibitem[{Saha et~al.(2023)Saha, Garimella, Kalyan, Pandey, Meher, Mathew, and
  Mukherjee}]{Saha2023}
Saha, P.; Garimella, K.; Kalyan, N.~K.; Pandey, S.~K.; Meher, P.~M.; Mathew,
  B.; and Mukherjee, A. 2023.
\newblock On the rise of fear speech in online social media.
\newblock \emph{PNAS}, 120(11): e2212270120.

\bibitem[{Samore et~al.(2018)Samore, Fessler, Holbrook, and
  Sparks}]{Samore2018}
Samore, T.; Fessler, D. M.~T.; Holbrook, C.; and Sparks, A.~M. 2018.
\newblock Electoral fortunes reverse, mindsets do not.
\newblock \emph{PLOS ONE}, 13(12): 1--15.

\bibitem[{Soroka, Fournier, and Nir(2019)}]{Soroka2019}
Soroka, S.; Fournier, P.; and Nir, L. 2019.
\newblock Cross-national evidence of a negativity bias in psychophysiological
  reactions to news.
\newblock \emph{PNAS}, 116(38): 18888--18892.

\bibitem[{Stone, Dunphy, and Smith(1966)}]{stone1966general}
Stone, P.~J.; Dunphy, D.~C.; and Smith, M.~S. 1966.
\newblock \emph{The general inquirer: A computer approach to content analysis.}
\newblock MIT press.

\bibitem[{Vaswani et~al.(2017)Vaswani, Shazeer, Parmar, Uszkoreit, Jones,
  Gomez, Kaiser, and Polosukhin}]{vaswani2017attention}
Vaswani, A.; Shazeer, N.; Parmar, N.; Uszkoreit, J.; Jones, L.; Gomez, A.~N.;
  Kaiser, {\L}.; and Polosukhin, I. 2017.
\newblock Attention is all you need.
\newblock \emph{NeurIPS}, 30.

\bibitem[{Veselovsky, Ribeiro, and West(2023)}]{veselovsky2023artificial}
Veselovsky, V.; Ribeiro, M.~H.; and West, R. 2023.
\newblock Artificial artificial artificial intelligence: Crowd workers widely
  use large language models for text production tasks.
\newblock \emph{arXiv preprint arXiv:2306.07899}.

\bibitem[{Vosoughi, Roy, and Aral(2018)}]{vosoughi2018spread}
Vosoughi, S.; Roy, D.; and Aral, S. 2018.
\newblock The spread of true and false news online.
\newblock \emph{Science}, 359(6380): 1146--1151.

\bibitem[{Wei et~al.(2022)Wei, Wang, Schuurmans, Bosma, Xia, Chi, Le, Zhou
  et~al.}]{wei2022chain}
Wei, J.; Wang, X.; Schuurmans, D.; Bosma, M.; Xia, F.; Chi, E.; Le, Q.~V.;
  Zhou, D.; et~al. 2022.
\newblock Chain-of-thought prompting elicits reasoning in large language
  models.
\newblock \emph{NeurIPS}, 35: 24824--24837.

\bibitem[{Youngblood et~al.(2023)Youngblood, Stubbersfield, Morin, Glassman,
  and Acerbi}]{Youngblood2023}
Youngblood, M.; Stubbersfield, J.~M.; Morin, O.; Glassman, R.; and Acerbi, A.
  2023.
\newblock Negativity bias in the spread of voter fraud conspiracy theory tweets
  during the 2020 US election.
\newblock \emph{Humanit. soc. sci.}, 10(1): 573.

\end{thebibliography}

\subsection{Paper Checklist}
\begin{enumerate}

\item For most authors...
\begin{enumerate}

    \item  Would answering this research question advance science without violating social contracts, such as violating privacy norms, perpetuating unfair profiling, exacerbating the socio-economic divide, or implying disrespect to societies or cultures?
    \answerYes{Yes.}
  \item Do your main claims in the abstract and introduction accurately reflect the paper's contributions and scope?
   Yes.
   \item Do you clarify how the proposed methodological approach is appropriate for the claims made? 
   \answerYes{Yes.}
   \item Do you clarify what are possible artifacts in the data used, given population-specific distributions?
    \answerYes{Yes, see Methods.}
  \item Did you describe the limitations of your work?
    \answerYes{Yes, see Limitations.}
  \item Did you discuss any potential negative societal impacts of your work?
    \answerYes{Yes, see Ethical Considerations.}
      \item Did you discuss any potential misuse of your work?
    \answerYes{Yes, see Ethical Considerations.}
    \item Did you describe steps taken to prevent or mitigate potential negative outcomes of the research, such as data and model documentation, data anonymization, responsible release, access control, and the reproducibility of findings?
    \answerYes{Yes, see Ethical considerations, as well as documentation/code in the code repository, discussions of anonymization in the Research Methods section.}
  \item Have you read the ethics review guidelines and ensured that your paper conforms to them?
   \answerYes{Yes.}
\end{enumerate}
\item Additionally, if your study involves hypotheses testing...

\begin{enumerate}
  \item Did you clearly state the assumptions underlying all theoretical results?
     \answerNA{NA}
  \item Have you provided justifications for all theoretical results?
   \answerNA{NA}
  \item Did you discuss competing hypotheses or theories that might challenge or complement your theoretical results?
  \answerNA{NA}
  \item Have you considered alternative mechanisms or explanations that might account for the same outcomes observed in your study?
   \answerNA{NA}
  \item Did you address potential biases or limitations in your theoretical framework?
   \answerNA{NA}
  \item Have you related your theoretical results to the existing literature in social science?
  \answerNA{NA}
  \item Did you discuss the implications of your theoretical results for policy, practice, or further research in the social science domain?
 \answerNA{NA}
\end{enumerate}
\item Additionally, if you are including theoretical proofs...

\begin{enumerate}
  \item Did you state the full set of assumptions of all theoretical results?
 \answerNA{NA}
	\item Did you include complete proofs of all theoretical results?
 \answerNA{NA}
\end{enumerate}
\item Additionally, if you ran machine learning experiments...

\begin{enumerate}
  \item Did you include the code, data, and instructions needed to reproduce the main experimental results (either in the supplemental material or as a URL)?
    \answerYes{Yes, see the Introduction for data and code.}
  \item Did you specify all the training details (e.g., data splits, hyperparameters, how they were chosen)?
    \answerYes{Yes, and further details are within the attached code.}
     \item Did you report error bars (e.g., with respect to the random seed after running experiments multiple times)?
    \answerYes{Yes}
	\item Did you include the total amount of compute and the type of resources used (e.g., type of GPUs, internal cluster, or cloud provider)?
    \answerYes{Yes, see subsection Model Training in the Research Methods section.}
     \item Do you justify how the proposed evaluation is sufficient and appropriate to the claims made? 
    \answerYes{Yes, see Methods.}
     \item Do you discuss what is ``the cost`` of misclassification and fault (in)tolerance?
    \answerYes{Yes, see Ethical Considerations.}
\end{enumerate}
\item Additionally, if you are using existing assets (e.g., code, data, models) or curating/releasing new assets, \textbf{without compromising anonymity}...
\begin{enumerate}
  \item If your work uses existing assets, did you cite the creators?
   \answerYes{Yes}
  \item Did you mention the license of the assets?
    \answerYes{Yes, the code is MIT licensed, as described in the repository link.}
  \item Did you include any new assets in the supplemental material or as a URL?
    \answerYes{Yes, we share code and data in our anonymized URL.}
  \item Did you discuss whether and how consent was obtained from people whose data you're using/curating?
   \answerYes{
   Yes, data was gathered without subject's consent as data are publicly avaialble and anonymized to remove any PII.}
  \item Did you discuss whether the data you are using/curating contains personally identifiable information or offensive content?
    \answerYes{Yes, no data contains PII.}
\item If you are curating or releasing new datasets, did you discuss how you intend to make your datasets FAIR (see \citet{fair})?
\answerYes{Yes, see Research Methods, page 4.}
\item If you are curating or releasing new datasets, did you create a Datasheet for the Dataset (see \citet{gebru2021datasheets})? 
\answerYes{Yes, see the code repository link.}
\end{enumerate}
\item Additionally, if you used crowdsourcing or conducted research with human subjects, \textbf{without compromising anonymity}...
\begin{enumerate}
  \item Did you include the full text of instructions given to participants and screenshots?
     \answerYes{Yes, see repository link.}
  \item Did you describe any potential participant risks, with mentions of Institutional Review Board (IRB) approvals?
    \answerYes{Yes, see Ethical Considerations.}
  \item Did you include the estimated hourly wage paid to participants and the total amount spent on participant compensation?
    \answerYes{Yes, see Methods.}
   \item Did you discuss how data is stored, shared, and deidentified?
  \answerYes{Yes, see Research Methods page 4.}
\end{enumerate}
\end{enumerate}

\newpage

\appendix
\renewcommand{\theequation}{S\arabic{equation}}
\renewcommand{\thetable}{S\arabic{table}}
\renewcommand{\thefigure}{S\arabic{figure}}
\setcounter{equation}{0}
\setcounter{table}{0}
\setcounter{figure}{0}

\section{Appendix}
\begin{figure}[tbh!]
    \centering
    \includegraphics[width=0.9\columnwidth]{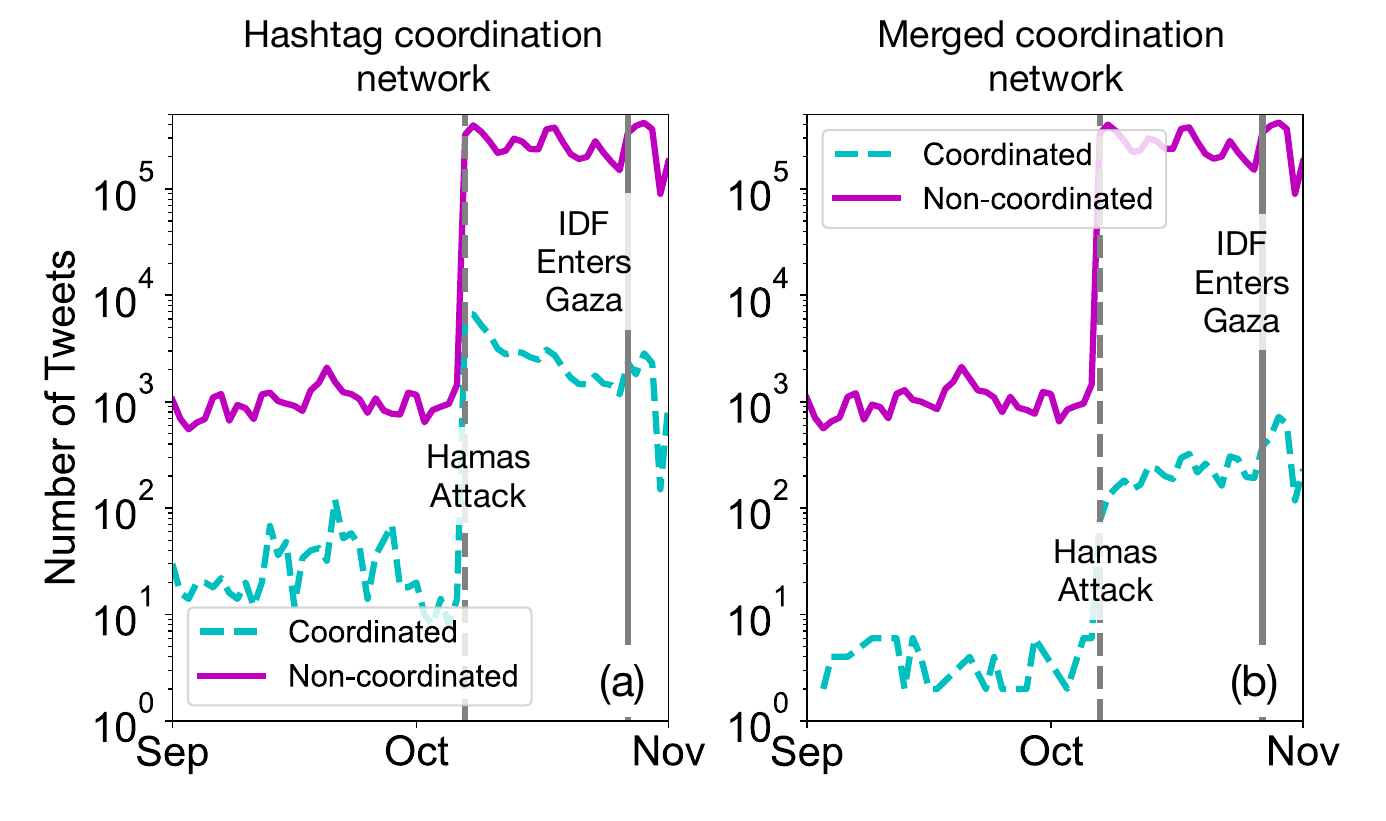}
    \caption{Post frequency over time for the Israel-Hamas dataset. Authentic and inauthentic coordinated accounts based on (a) hashtags (main text), and (b) merged similarity networks \cite{luceri2023unmasking}. Vertical lines correspond to the Hamas attack on October 7th, 2023, and the Israel Defense Force (IDF) entering Gaza on October 27th.}
    \label{fig:hamas-tweet-freq}
\end{figure}

\begin{table*}[tbh!]
    \centering
   \footnotesize
\begin{tabular}{|l|p{7cm}|p{2.7cm}|}
\hline
    \textbf{Dataset} & \textbf{Text} & \textbf{Ground truth label} \\ \hline
    Random posts & Aircraft crashes into residential building in Russian city near Crimea, killing at least three https://xxxx & {\color{red} Hazard} \\ \hline
    Random posts & @XXXX @YYYY But he is a liar and was thrown out because he is a liar, we know you love him but really you have to be aware that he is a Liar x & {\color{olive} No hazard} \\ \hline
    2023 Israel-Hamas War & @XX Just surreal! Footage of Palestinian Hamas terrorists who infiltrated into Israel from Gaza, firing at residents in Sderot from an SUV.& {\color{red} Hazard} \\ \hline
    2023 Israel-Hamas War &@XX Saudi Arabia’s NEW official map scraps Israel and only names Palestine  & {\color{olive} No hazard} \\ \hline
    2022 French Election &[translated from French] @XX Macronia panics and talks about a Republican front against the left. No hesitation, let us block to their Republic the possessors, social inequalities, that of repression and police violence, for a front of anti-liberal and even anti-capitalist struggles.  & {\color{red} Hazard} \\ \hline
    2022 French Election &[translated from French] Among the breakers of \#1May2022 are the sbores of \#Macron such as Roland Branleur alias @tarzoondtc troublemaker a seasoned disciple of Benala dt the objective is to decredivize the movements of struggle against the macroistic oligarchy. We wish him good health  & {\color{olive} No hazard} \\ \hline    
    \textbf{Dataset} & \textbf{Text} & \textbf{Hazard~model}\\ &&\textbf{confidence} \\ \hline
        2023 Israel-Hamas War & RT @XX We’ll be streaming here on Monday from 1pm ET, don’t miss it  &  {\color{olive} 0.05} \\ \hline
    2023 Israel-Hamas War & @XX Israeli airstrikes flattened mosques over the heads of worshipers. At least 2 hospitals and 2 centers run by Palestine Red Crescent Society, have been hit. So have two schools run by the U.N. agency Israel fighter jets/artillery have struck targets in Gaza frequently over years &  {\color{red} 0.92} \\ \hline
   2022 French Election & [translated from French] Is there any hope that you will have 500 referrals by Monday?& {\color{olive}0.03}\\\hline
   2022 French Election & @XX Putin is a murderer more dangerous than Osama bin Laden. Threats the world with a nuclear bomb. He, like no one else, can just press the red button. Gotta get him dead or alive. His immediate surroundings have the most opportunities for this. Otherwise he'll kill us. & {\color{red} 0.94}\\\hline
\end{tabular}
    \caption{Hazard ground truth labels and model prediction examples.}
    \label{tab:hazard_ex}
\end{table*}

\begin{table*}[tbh!]
    \centering
    \footnotesize
    \begin{tabular}{|l|p{2cm}|p{2cm}|p{2cm}|p{2cm}|}
    \hline
\textbf{Rank} & \textbf{Highest hazard word} &\textbf{Highest hazard ratio}&  \textbf{Lowest hazard word}&\textbf{Lowest hazard ratio}\\\hline\hline
1&mask&257.513&friday&0.008\\\hline
2&residential&218.82&historical&0.01\\\hline
3&babies&206.96&bible&0.01\\\hline
4&children&205.286&jesus&0.012\\\hline
5&buildings&170.045&davido&0.013\\\hline
6&[translated from Arabic] children&169.452&roman&0.015\\\hline
7&taxes&155.664&episode&0.016\\\hline
8&[translated from Arabic] hospital&149.883&labor&0.017\\\hline
9&[translated from Arabic] hospital&132.76&messi&0.019\\\hline
10&woodward&132.092&rally&0.019\\\hline
\end{tabular}
    \caption{Words associated with high and low-hazard posts in the full Israel-Hamas war dataset. Words have a minimum frequency of $10^{-6}$ in each set of posts.}
    \label{tab:israel_hazards}
\end{table*}

\begin{table*}[tbh!]
    \centering
    \footnotesize
    \begin{tabular}{|l|p{2cm}|p{2cm}|p{2cm}|p{2cm}|}
    \hline
\textbf{Rank} & \textbf{Highest hazard word} &\textbf{Highest hazard ratio}&  \textbf{Lowest hazard word}&\textbf{Lowest hazard ratio}\\\hline\hline
1&children&287.486&parodique&0.011\\\hline
2&war&272.263&18h&0.015\\\hline
3&ukrainian&180.407&film&0.018\\\hline
4&policiers&176.809&20h&0.021\\\hline
5&sanitaire&172.954&netflix&0.023\\\hline
6&régime&169.956&youtube&0.025\\\hline
7&weapons&169.202&épisode&0.025\\\hline
8&dictature&166.016&impatience&0.025\\\hline
9&dangereux&146.117&émission&0.027\\\hline
10&nuclear&140.831&rendez-vous&0.027\\\hline\hline
\end{tabular}
    \caption{Words associated with high and low-hazard posts in the full 2022 French Election dataset. Words have a minimum frequency of $10^{-6}$ in each set of posts.}
    \label{tab:phase1b_hazards}
\end{table*}

\begin{figure}[tbh!]
    \centering
    \includegraphics[width=0.9\columnwidth]{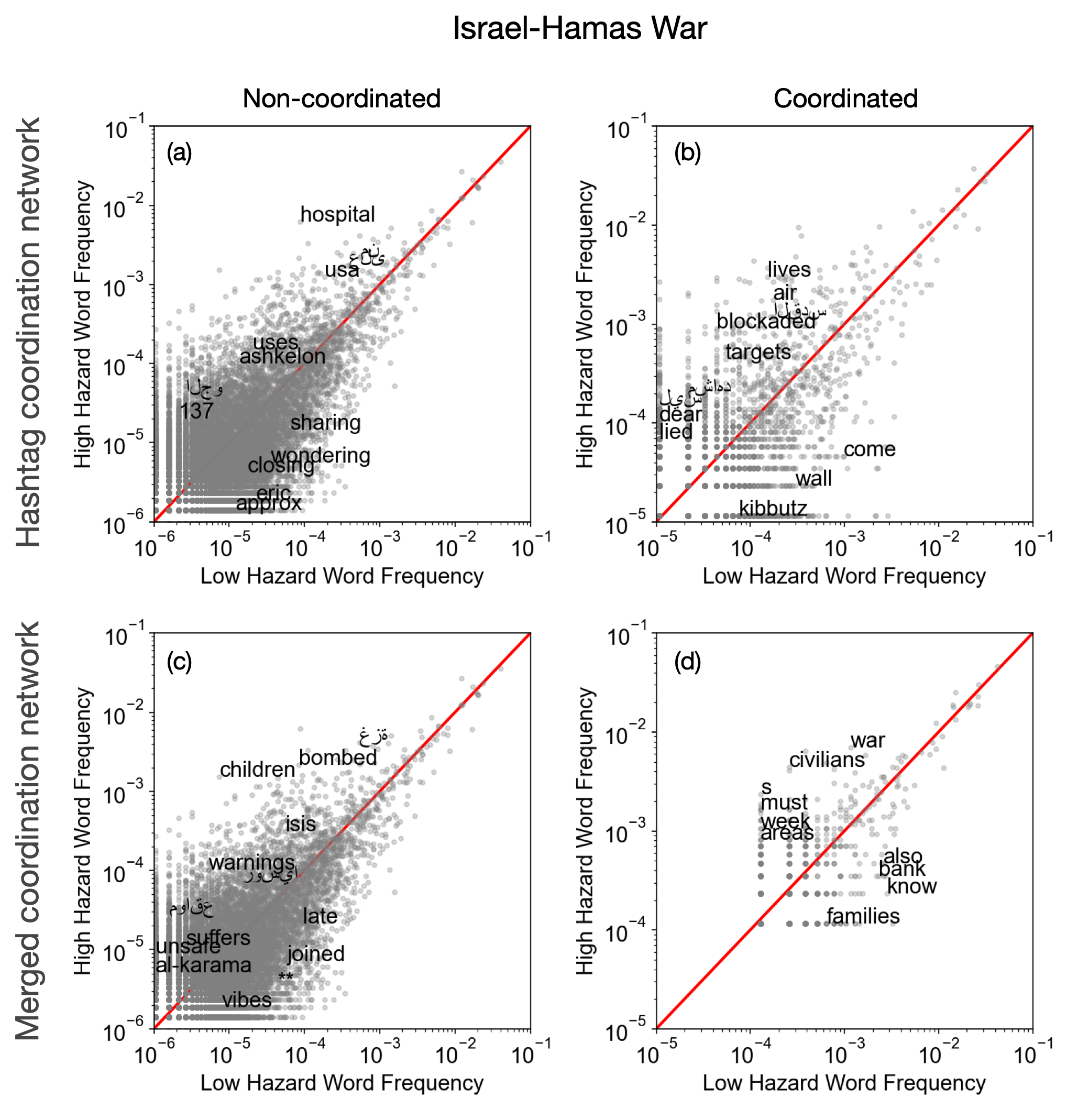}
    \caption{Words associated with high-confidence and low-confidence hazard posts among coordinated and authentic accounts within the Israel-Hamas war dataset. (a--b) Hashtag-based coordination networks, (c--d) merged coordination networks \cite{luceri2023unmasking}.}
\label{fig:haz_freq_coord-ih}
\end{figure}

\begin{figure}[tbh!]
    \centering
    \includegraphics[width=0.9\columnwidth]{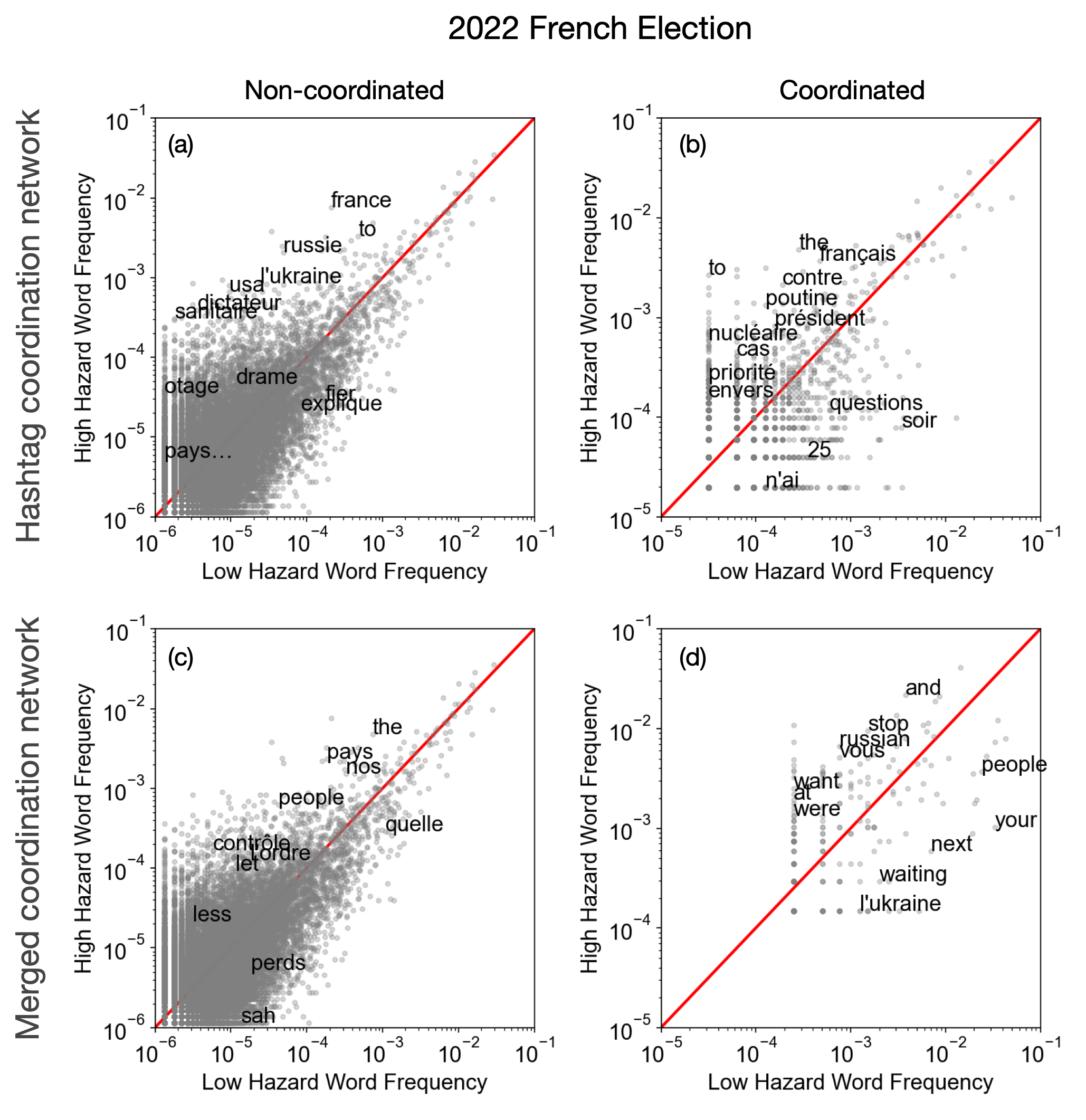}
    \caption{Words associated with high-confidence and low-confidence hazard posts among coordinated and authentic accounts within the 2022 French Election dataset. (a--b) Hashtag-based coordination networks, (c--d) merged coordination networks \cite{luceri2023unmasking}.}
\label{fig:haz_freq_coord-french}
\end{figure}

\begin{table*}[tbh!]
    \centering
    \footnotesize
    \begin{tabular}{|l|p{2cm}|p{2cm}|p{2cm}|p{2cm}|}
    \hline
\textbf{Rank} & \textbf{Highest hazard word} &\textbf{Highest hazard ratio}&  \textbf{Lowest hazard word}&\textbf{Lowest hazard ratio}\\\hline\hline
1&openrafahcrossing&248.099&london&0.004\\\hline
2&christanhospitalingaza&163.98&mia&0.005\\\hline
3&communications&95.3&khalifa&0.005\\\hline
4&warplanes&88.379&•&0.011\\\hline
5&knew&83.055&border&0.015\\\hline
6&green&81.99&received&0.016\\\hline
7&starlink&80.925&friend&0.018\\\hline
8&aircraft&71.342&manipulate&0.018\\\hline
9&building&68.147&[translated from Arabic] Israel&0.019\\\hline
10&that\'s&67.881&show&0.021\\\hline
\end{tabular}
    \caption{Words associated with high and low-hazard posts among coordinated accounts in the Israel-Hamas war dataset based on hashtag coordination networks. Words have a minimum frequency of $10^{-5}$ in each set of posts.}
    \label{tab:challenge_hazards_coord_hash}
\end{table*}

\begin{table*}[tbh!]
    \centering
    \footnotesize
\begin{tabular}{|l|p{2cm}|p{2cm}|p{2cm}|p{2cm}|}
    \hline
\textbf{Rank} & \textbf{Highest hazard word} &\textbf{Highest hazard ratio}&  \textbf{Lowest hazard word}&\textbf{Lowest hazard ratio}\\\hline\hline
1&s&18.137&history&0.036\\\hline
2&civilians&17.23&know&0.082\\\hline
3&hospital&16.626&silent&0.082\\\hline
4&bombed&15.87&remain&0.091\\\hline
5&hospitals&14.51&feel&0.091\\\hline
6&airstrikes&13.603&through&0.113\\\hline
7&must&12.696&between&0.113\\\hline
8&press&12.696&west&0.13\\\hline
9&year&12.696&said&0.13\\\hline
10&killing&11.789&happening&0.13\\\hline
\end{tabular}
    \caption{Words associated with high and low-hazard posts among coordinated accounts in the Israel-Hamas war dataset based on merged coordination networks \cite{luceri2023unmasking}. Words have a minimum frequency of $10^{-5}$ in each set of posts.}
    \label{tab:challenge_hazards_coord_merged}
\end{table*}

\begin{table*}[tbh!]
    \centering
    \footnotesize
\begin{tabular}{|l|p{2cm}|p{2cm}|p{2cm}|p{2cm}|}
    \hline
\textbf{Rank} & \textbf{Highest hazard word} &\textbf{Highest hazard ratio}&  \textbf{Lowest hazard word}&\textbf{Lowest hazard ratio}\\\hline\hline
1&mask&220.641&friday&0.008\\\hline
2&residential&207.869&bible&0.01\\\hline
3&babies&206.401&jesus&0.012\\\hline
4&buildings&196.066&historical&0.012\\\hline
5&children&188.24&roman&0.015\\\hline
6&[translated from Arabic] children&161.627&messi&0.015\\\hline
7&taxes&146.653&davido&0.016\\\hline
8&[translated from Arabic] hospital&133.148&episode&0.016\\\hline
9&woodward&127.276&labour&0.017\\\hline
10&gazahospitalbombing&126.395&recognized&0.019\\\hline
\end{tabular}
    \caption{Words associated with high and low-hazard posts among authentic accounts in the Israel-Hamas war dataset based on hashtag coordination networks. Words have a minimum frequency of $10^{-6}$ in each set of posts.}
    \label{tab:challenge_hazards_noncoord_hash}
\end{table*}

\begin{table*}[tbh!]
    \centering
    \footnotesize
\begin{tabular}{|l|p{2cm}|p{2cm}|p{2cm}|p{2cm}|}
    \hline
\textbf{Rank} & \textbf{Highest hazard word} &\textbf{Highest hazard ratio}&  \textbf{Lowest hazard word}&\textbf{Lowest hazard ratio}\\\hline\hline
1&mask&257.447&friday&0.008\\\hline
2&residential&218.541&historical&0.01\\\hline
3&babies&206.61&bible&0.01\\\hline
4&children&205.17&jesus&0.012\\\hline
5&buildings&169.853&davido&0.013\\\hline
6&[translated from Arabic] children&169.408&roman&0.015\\\hline
7&taxes&155.624&episode&0.016\\\hline
8&[translated from Arabic] hospital&149.844&labor&0.017\\\hline
9&[translated from Arabic] hospital&132.725&rally&0.019\\\hline
10&woodward&132.058&christ&0.019\\\hline
\end{tabular}
    \caption{Words associated with high and low-hazard posts among authentic accounts in the Israel-Hamas war dataset based on merged coordination networks \cite{luceri2023unmasking}. Words have a minimum frequency of $10^{-6}$ in each set of posts.}
    \label{tab:challenge_hazards_noncoord_merged}
\end{table*}

\begin{table*}[tbh!]
    \centering
    \footnotesize
    \begin{tabular}{|l|p{2cm}|p{2cm}|p{2cm}|p{2cm}|}
    \hline
\textbf{Rank} & \textbf{Highest hazard word} &\textbf{Highest hazard ratio}&  \textbf{Lowest hazard word}&\textbf{Lowest hazard ratio}\\\hline\hline
1&to&84.905&rendez-vous&0.006\\\hline
2&are&63.521&jevotezemmour&0.008\\\hline
3&\&&54.087&meeting&0.008\\\hline
4&guerre&48.742&demain&0.008\\\hline
5&of&40.88&mars&0.017\\\hline
6&russie&35.849&direct&0.021\\\hline
7&enfants&30.817&hashtag&0.021\\\hline
8&sky&27.673&pouvez&0.022\\\hline
9&ukraine&25.0&soir&0.023\\\hline
10&we&24.528&[Fist emoji]&0.023\\\hline
\end{tabular}
    \caption{Words associated with high and low-hazard posts among coordinated accounts in the 2022 French Election dataset based on hashtag coordination networks. Words have a minimum frequency of $10^{-5}$ in each set of posts.}
    \label{tab:phase1b_hazards_coord_hash}
\end{table*}
\begin{table*}[tbh!]
    \centering
    \footnotesize
    \begin{tabular}{|l|p{2cm}|p{2cm}|p{2cm}|p{2cm}|}
    \hline
\textbf{Rank} & \textbf{Highest hazard word} &\textbf{Highest hazard ratio}&  \textbf{Lowest hazard word}&\textbf{Lowest hazard ratio}\\\hline\hline
1&our&42.859&future&0.028\\\hline
2&children&29.355&your&0.031\\\hline
3&on&19.375&where&0.045\\\hline
4&killed&17.026&than&0.046\\\hline
5&missiles&15.265&rt&0.059\\\hline
6&these&14.678&next&0.084\\\hline
7&poutine&12.916&can&0.087\\\hline
8&want&9.981&do&0.09\\\hline
9&russian&8.807&what&0.098\\\hline
10&defense&8.807&nous&0.098\\\hline
\end{tabular}
    \caption{Words associated with high and low-hazard posts among coordinated accounts in the 2022 French Election dataset based on merged coordination networks \cite{luceri2023unmasking}. Words have a minimum frequency of $10^{-5}$ in each set of posts.}
    \label{tab:phase1b_hazards_coord_merged}
\end{table*}

\begin{table*}[tbh!]
    \centering
    \footnotesize
    \begin{tabular}{|l|p{2cm}|p{2cm}|p{2cm}|p{2cm}|}
    \hline
\textbf{Rank} & \textbf{Highest hazard word} &\textbf{Highest hazard ratio}&  \textbf{Lowest hazard word}&\textbf{Lowest hazard ratio}\\\hline\hline
1&war&270.952&parodique&0.011\\\hline
2&ukrainian&178.578&18h&0.016\\\hline
3&policiers&175.836&film&0.016\\\hline
4&sanitaire&172.366&20h&0.021\\\hline
5&régime&169.666&netflix&0.023\\\hline
6&weapons&166.993&hashtag&0.024\\\hline
7&dictature&165.425&youtube&0.025\\\hline
8&dangereux&146.016&épisode&0.025\\\hline
9&sky&139.435&impatience&0.026\\\hline
10&nuclear&136.076&émission&0.027\\\hline
\end{tabular}
    \caption{Words associated with high and low-hazard posts among authentic accounts in the 2022 French Election dataset based on hashtag coordination networks. Words have a minimum frequency of $10^{-6}$ in each set of posts.}
    \label{tab:phase1b_hazards_noncoord_hash}
\end{table*}

\begin{table*}[tbh!]
    \centering
    \footnotesize
    \begin{tabular}{|l|p{2cm}|p{2cm}|p{2cm}|p{2cm}|}
    \hline
\textbf{Rank} & \textbf{Highest hazard word} &\textbf{Highest hazard ratio}&  \textbf{Lowest hazard word}&\textbf{Lowest hazard ratio}\\\hline\hline
1&children&270.12&parodique&0.011\\\hline
2&war&266.496&18h&0.015\\\hline
3&policiers&177.052&film&0.018\\\hline
4&sanitaire&173.024&20h&0.021\\\hline
5&régime&170.368&netflix&0.023\\\hline
6&ukrainian&168.139&youtube&0.025\\\hline
7&dictature&166.34&épisode&0.025\\\hline
8&weapons&162.792&impatience&0.025\\\hline
9&dangereux&146.397&émission&0.027\\\hline
10&criminel&128.89&rendez-vous&0.027\\\hline
\end{tabular}
    \caption{Words associated with high and low-hazard posts among authentic accounts in the 2022 French Election dataset based on merged coordination networks \cite{luceri2023unmasking}. Words have a minimum frequency of $10^{-6}$ in each set of posts.}
    \label{tab:phase1b_hazards_noncoord_merged}
\end{table*}

\begin{figure*}[tbh!]
    \centering
    \includegraphics[width=0.7\linewidth]{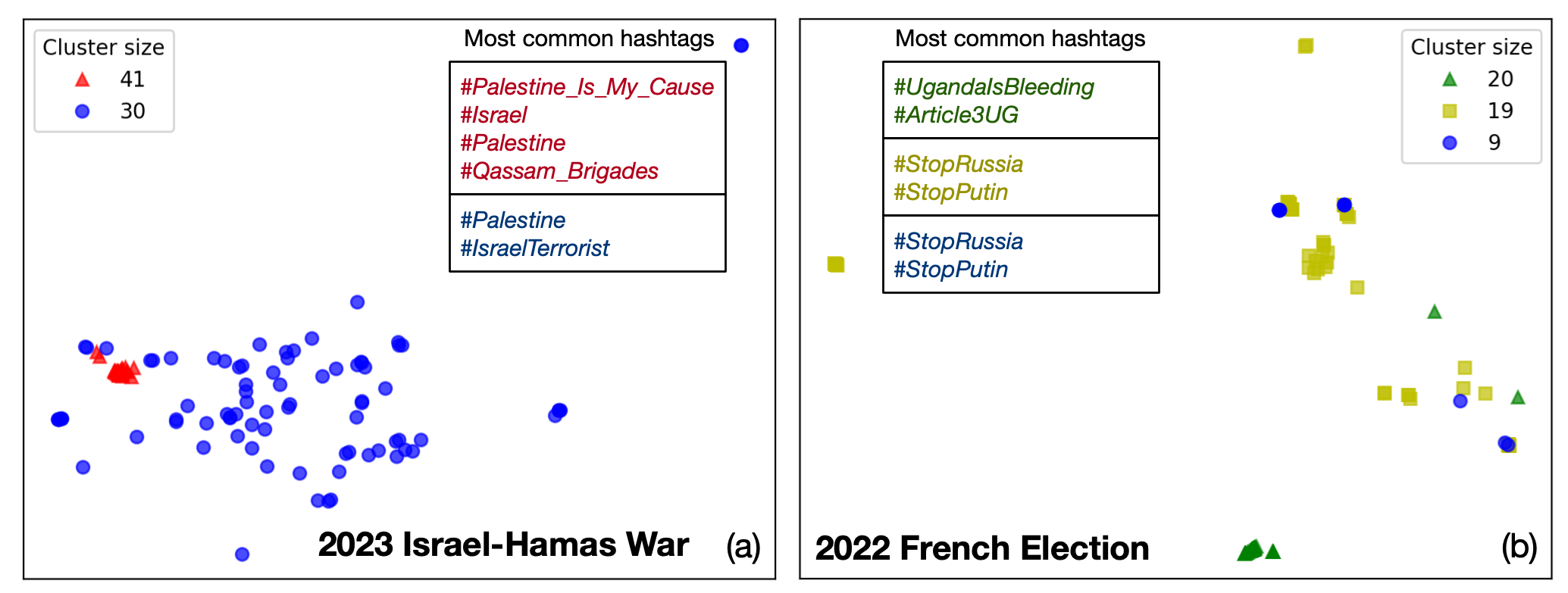}
    \caption{Embeddings text from clusters shown in in Fig.~\ref{fig:coord-hazard}. (a) The 2023 Israel-Hamas war dataset, and (b) the 2022 French election dataset. Plots are colored by the cluster size, while the most popular hashtags' fonts are colored based on the cluster color. We subsample 5000 posts from each set of coordinated accounts found from the hashtag-based coordination networks, then embed the text using \texttt{distiluse-base-multilingual-cased-v2} (the same text embedding model used when detecting hazards). These embeddings are then compressed to two dimensions using UMAP \cite{mcinnes2018umap}. For clarity, we only show the posts associated with the posts shown in Fig.~\ref{fig:coord-hazard} (121 and 147 for the respective subfigures). }
    \label{fig:hazard_embedding}
\end{figure*}

\begin{figure}[tbh!]
    \centering
    \includegraphics[width=0.9\columnwidth]{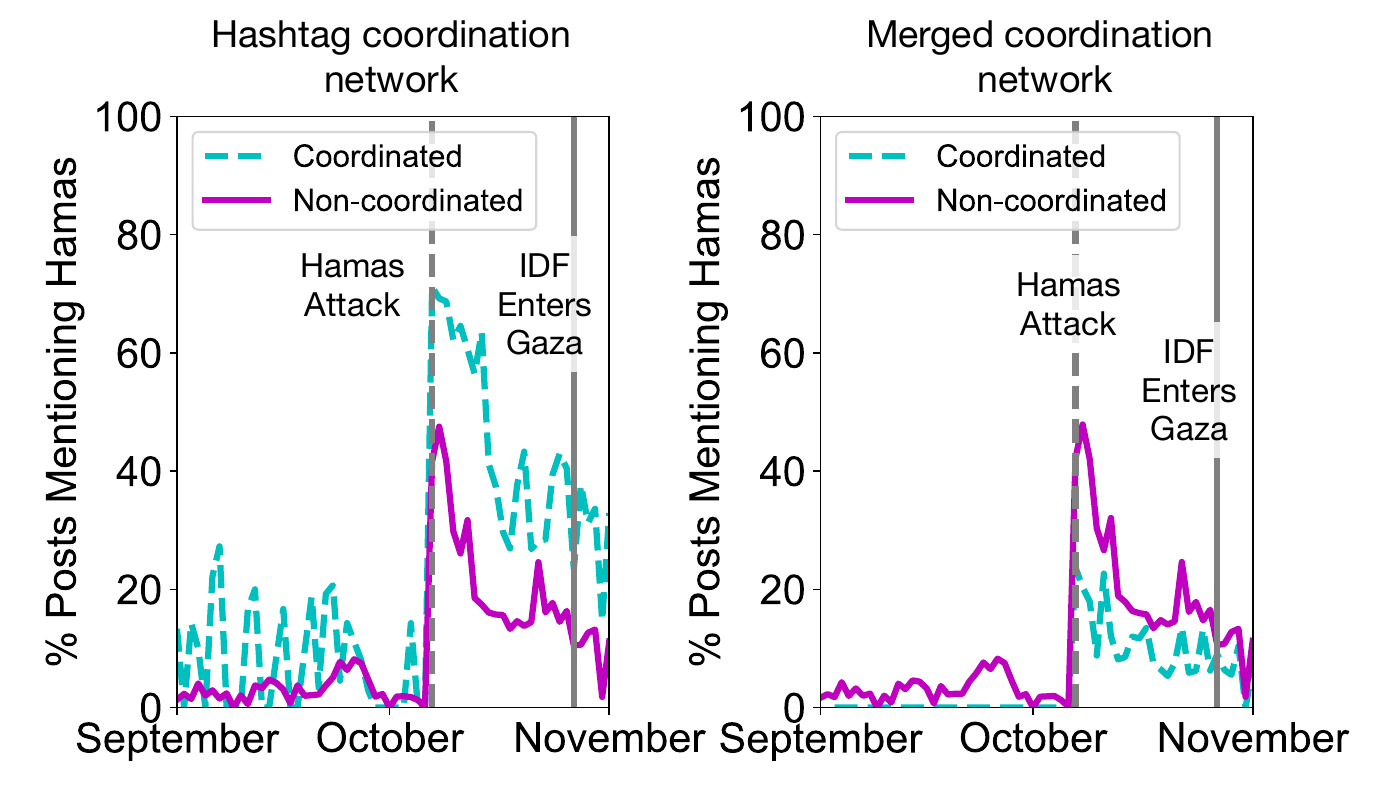}
    \caption{Percent of posts mentioning Hamas over time for (a) authentic, and (b) inauthentic coordinated accounts. Vertical lines correspond to the Hamas attack on October 7th, 2023, and the Israel Defense Force (IDF) entering Gaza on October 27th.}
    \label{fig:hamas-ment-freq}
\end{figure}
\begin{figure}[tbh!]
    \centering
    \includegraphics[width=0.9\columnwidth]{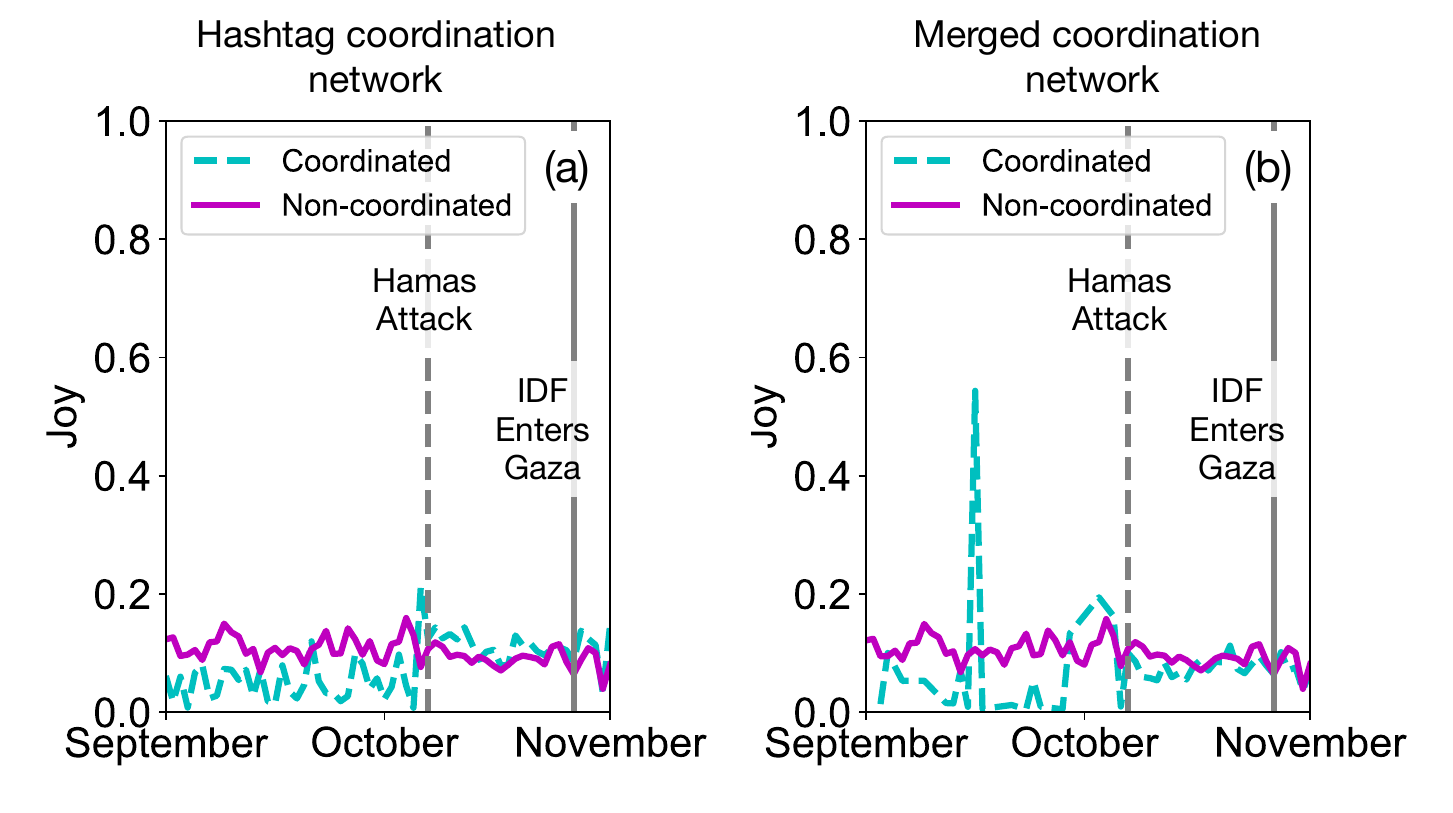}
    \caption{Mean joy emotion confidence over time for authentic and inauthentic coordinated accounts using Demux \cite{Chochlakis2023}. (a) Hashtag-based coordination network and (b) merged coordination network \cite{luceri2023unmasking}. Vertical lines correspond to the Hamas attack on October 7th, 2023, and the Israel Defense Force (IDF) entering Gaza on October 27th.}
    \label{fig:hamas-joy-freq}
\end{figure}
\begin{figure}[tbh!]
    \centering
    \includegraphics[width=0.9\columnwidth]{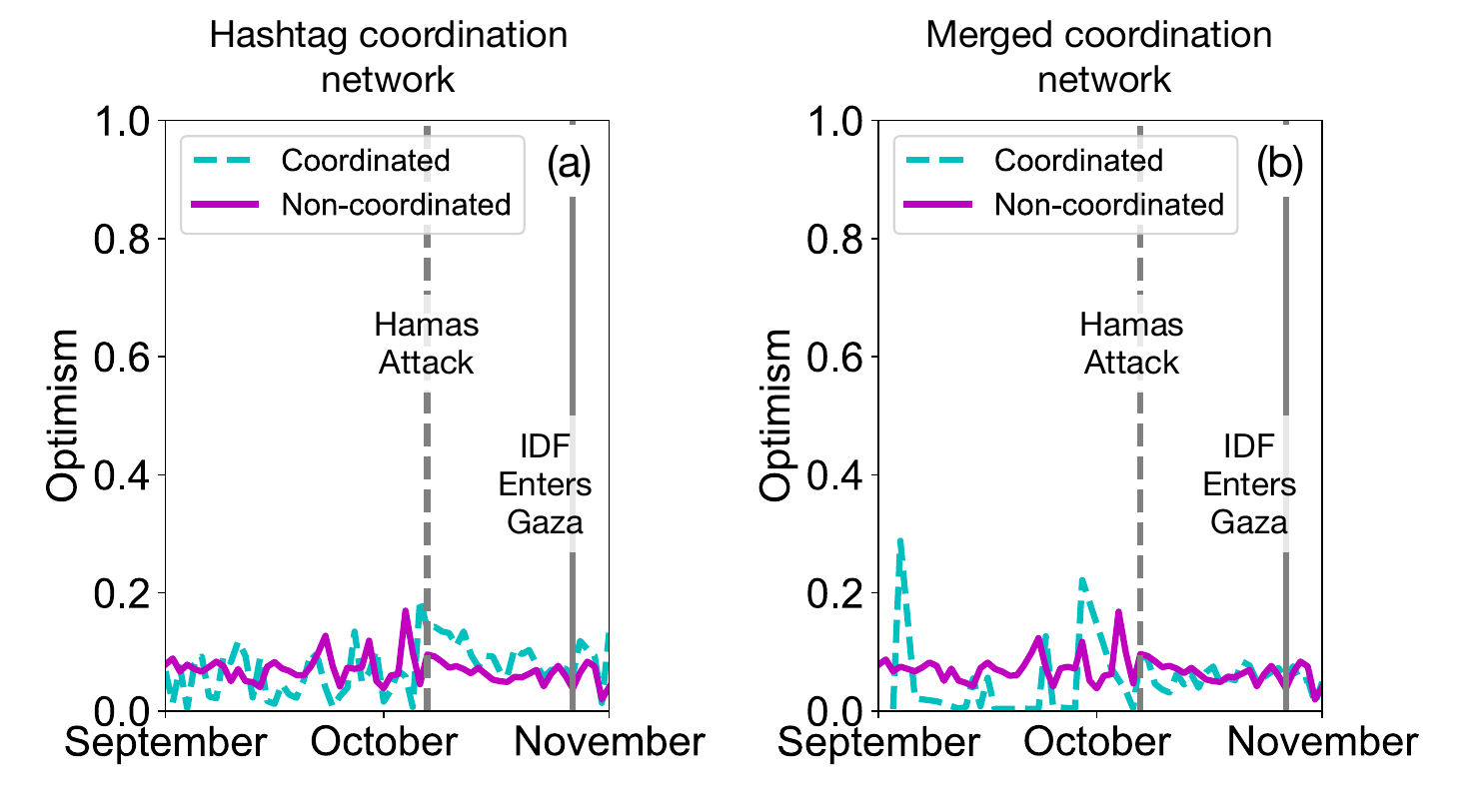}
    \caption{Mean optimism emotion confidence over time for authentic and inauthentic coordinated accounts using Demux \cite{Chochlakis2023}. (a) Hashtag-based coordination network and (b) merged coordination network \cite{luceri2023unmasking}. Vertical lines correspond to the Hamas attack on October 7th, 2023, and the Israel Defense Force (IDF) entering Gaza on October 27th.}
    \label{fig:hamas-opt-freq}
\end{figure}

\begin{figure}[tbh!]
    \centering
    \includegraphics[width=0.9\columnwidth]{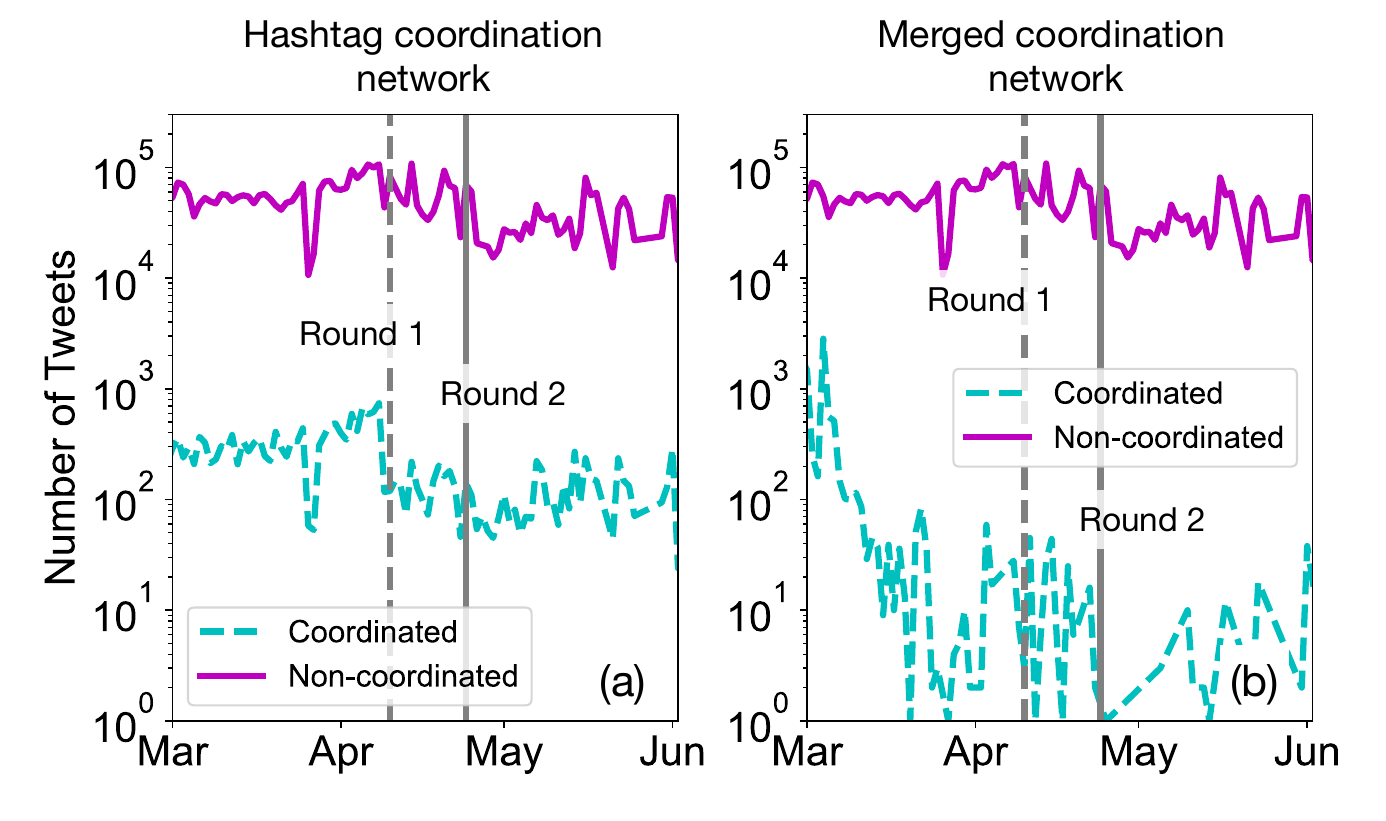}
    \caption{Post frequency over time for the 2022 French Election among authentic and inauthentic coordinated accounts. (a) Hashtag-based coordination network and (b) merged coordination network \cite{luceri2023unmasking}.}
    \label{fig:phase1b-tweet-freq}
\end{figure}

\begin{figure}[tbh!]
    \centering
    \includegraphics[width=0.9\columnwidth]{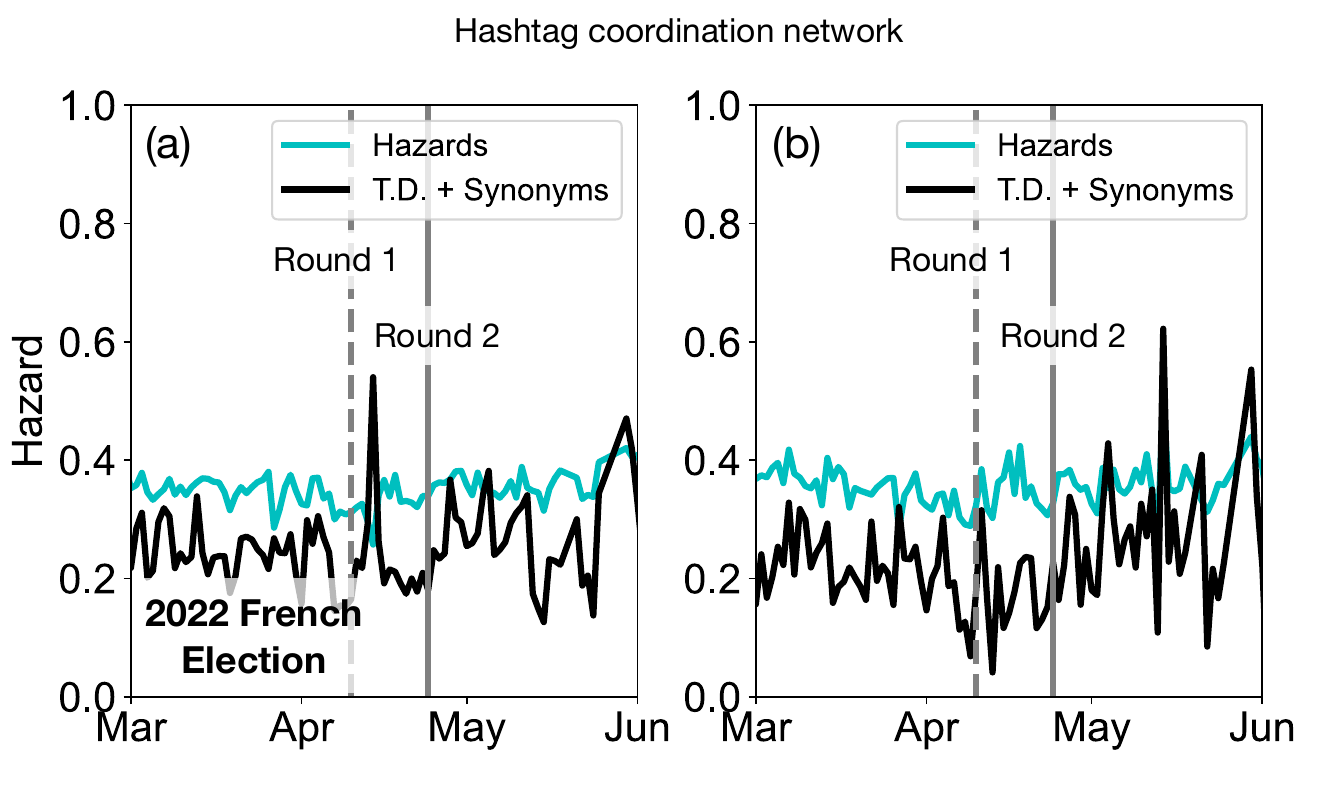}
    \caption{Hazards and threats over time for the 2022 French Election  dataset. The plots show mean hazard confidences each day as well as the overall mean proportion of posts with at least one word from the Threat Dictionary \cite{Choi2022} + Synonyms for authentic and inauthentic coordinated accounts. (a) Authentic account posts and (b) hashtag coordination network posts. Vertical lines correspond to Round 1 voting (April 10, 2022) and the Round 2 runoff (April 24, 2022).
 }
    \label{fig:phase1b-hazard_hash}
\end{figure}

\begin{figure}[tbh!]
    \centering
    \includegraphics[width=1\columnwidth]{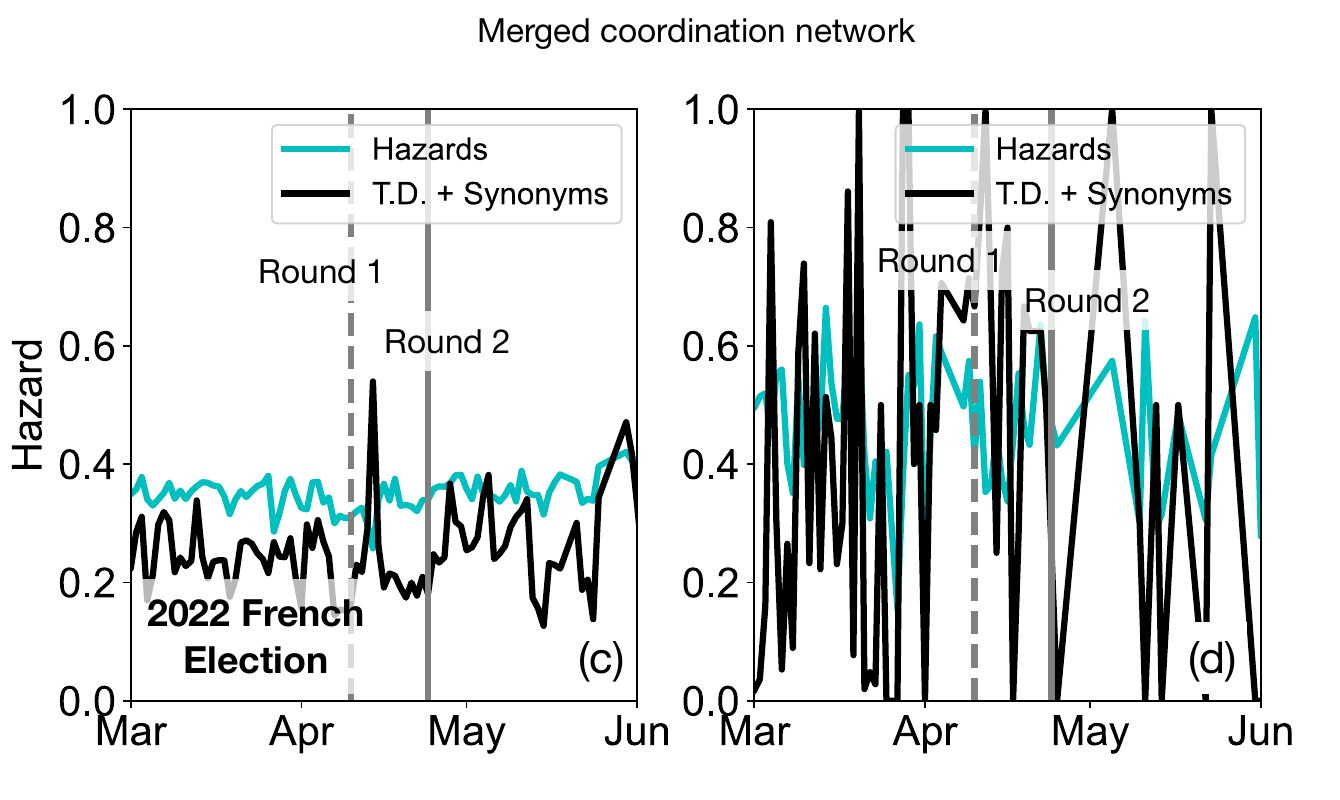}
    \caption{Hazards and threats over time for the 2022 French Election  dataset. The plots show mean hazard confidences each day as well as the overall mean proportion of posts with at least one word from the Threat Dictionary \cite{Choi2022} + Synonyms for (a) authentic accounts and (b) inauthentic coordinated accounts, where coordination is uncovered from merged coordination accounts \cite{luceri2023unmasking}. Vertical lines correspond to Round 1 voting (April 10, 2022) and the Round 2 runoff (April 24, 2022).
 }
    \label{fig:phase1b-hazard_merged}
\end{figure}

\end{document}